# IMPERIAL

**Mrs Ravinder Panesar**
Senior Postgraduate (MSc) Administrator

10/03/2025

To whom it may concern,

**Name:** KUMAR Swapnil
**CID:** 02146391

I confirm that the student named above has completed the MSc in Advanced Aeronautical Engineering in 2024. His Major Individual Research Project title is "Uncertainty Quantification for multifidelity simulations". The project was supervised by Professor Francesco Montomoli.

Yours faithfully,

**Ravinder Panesar**
**Senior Postgraduate (MSc) Administrator**

Imperial College London
Department of Aeronautics

# Uncertainty Quantification for Multi-Fidelity Simulations

Swapnil Kumar, Francesco Montomoli

# Acknowledgements

I would like thank my supervisor Professor, Francesco Montomoli for helping me to develop my research skills and to think critically about my work, and also for his guidance and supervision and for conducting more than 30+ meetings throughout my Master Thesis.

Furthermore, I am also grateful to Professor, Spencer Sherwin (Head of Department, Department of Aeronautics) and Professor, Professor Jason Riley (Vice-Dean, Imperial College London) for their consistent support throughout my graduate studies at Imperial College London. Their guidance and mentorship were invaluable to my success.

# Contents

vi

# List of Tables

# List of Figures

# Abstract

Uncertainty quantification strategies like Co-kriging are crucial in industrial application because of the fact, high-fidelity models are used for the final validation purpose. This research primarily focuses on gathering the high-fidelity and low-fidelity simulation data using Nektar++ and XFOIL package respectively. The utilization of the higher polynomial distribution in calculating the Coefficient of lift and drag has demonstrated superior accuracy and precision. Further, Co-kriging Data fusion and Adaptive sampling technique has been used to obtain the precise data predictions for the lift and drag within the confined domain (for the angle of attack ranging from 1 to 7 degree) without conducting the costly simulations on HPC clusters. This creates a methodology to quantifying uncertainty in computational fluid dynamics by minimizing the required number of samples [15]. To minimize the reliability on high-fidelity numerical simulations in Uncertainty Quantification, a multi-fidelity strategy has been adopted. The effectiveness of the multi-fidelity deep neural network model has been validated through the approximation of benchmark functions across 1-, 32-, and 100-dimensional, encompassing both linear and non-linear correlations [16]. The surrogate modelling results showed that multi-fidelity deep neural network model has shown excellent approximation capabilities for the test functions and multi-fidelity deep neural network method has outperformed Co-kriging in effectiveness [16]. In addition to that, multi-fidelity deep neural network model is utilized for the simulation of aleatory uncertainty propagation in 1-, 32-, and 100-dimensional function test, considering both uniform and Gaussian distributions for input uncertainties. The results have shown that multi-fidelity deep neural network model has efficiently predicted the probability density distributions of quantities of interest as well as the statistical moments with precision and accuracy [16]. The Co-Kriging model has exhibited limitations when addressing 32-Dimension problems due to the limitation of memory capacity for storage and manipulation [16]. In addition to that, Radial bias function, Kriging, and Co-Kriging models has possesses computational challenges for 100-Dimension function, because of the required system memory for training.

# Chapter 1

# Introduction

## 1.1 Background

Computational fluid dynamics (CFD) is a branch of fluid dynamics that uses numerical techniques to solve the Navier–Stokes equations (Incompressible and compressible), which govern the motion of fluids. CFD is a computationally intensive field, requiring large amounts of computing power and efficiency, and it is very important to the design of engineering systems and components (CFD evaluates the flow phenomena associated with the given system [3]). The prediction accuracy of the CFD simulations depends on the modelling capabilities and computational resources, and it is important to have a high level of confidence in the CFD results. Now a days, high-fidelity simulations are being preferred in the academia and industry due to its accuracy and high prediction.

In context of numerical simulation, High-fidelity simulation data refers to the simulations being conducted on HPC clusters [3] and it requires high GPU units. It is also referred as expensive data [15] in the context of Uncertainty Quantification. Low-fidelity simulation data refers to the simulations which do not requires high GPU units. It is also referred as cheap data [15] in the context of Uncertainty Quantification. High-fidelity and low-fidelity simulation are relative terms. In context of DNS/LES simulations, RANS simulation is a low-fidelity simulation.

Uncertainty quantification and computational fluid dynamics are important evaluations in the aerospace industry (The industry uses Uncertainty quantification models to predict the computational fluid dynamics data which are computationally expensive). The reliability of CFD simulations is increasingly affected by uncertainties as modeling and numerical errors are reduced (Nektar++ is a high-fidelity computational fluid dynamics (CFD) software that can provide more precise results than commercial CFD packages). It is also important to understand the uncertainty of adding robustness in the design process for gas turbines and combustion chambers in aircraft engines[20]

In Computational fluid dynamics and Uncertainty Quantification, one of the key goals is to assess how different factors and parameters affect the accuracy and reliability of CFD simulations. The lift coefficient versus angle of attack is a fundamental metric used in aerodynamics and fluid dynamics to evaluate the lift generated by an airfoil at varying angles of attack.

The Monte Carlo Method is a widely accepted approach for quantifying uncertainty, but its thorough application in Computational Fluid Dynamics evaluation is computationally expensive. Conversely, Surrogate models have gained prominence in uncertainty quantification due to their ability to achieve similar results with fewer Computational Fluid Dynamics evaluations compared to the Monte Carlo Method. Among these Surrogate models, Kriging stands out. Kriging utilizes prior distribution and known sample points to interpolate functional estimations of stochastic output. To enhance the efficiency of Kriging surrogates for achieving stochastic convergence or optimizing designs without significant computational overhead, Co-kriging comes into play by integrating lower fidelity data into the surrogate model. Additionally, adaptive sampling has emerged as a crucial technique for constructing surrogate models efficiently [15].

To further minimize the reliance on high-fidelity numerical simulations in Uncertainty Quantification, multi-fidelity strategy is used [17]. The main principle behind the multi-fidelity method is to maintain the accuracy of the surrogate model's predictions by leveraging a large volume of low-fidelity data complemented by a smaller set of high-fidelity data [17]

## 1.2 Literature Review

Commercial CFD software packages use a variety of numerical solvers, including finite difference, finite volume, and finite element solvers. Finite difference methods (FDMs) represent the most established numerical techniques for Computational Fluid Dynamics, involving the discretization of the governing equations in both space and time through a grid of points. Implementing FDMs is straightforward and highly efficient when dealing with uncomplicated geometries, but their application becomes challenging when addressing complex geometries, potentially leading to reduced accuracy compared to other methods (For more information, refer Appendix)

### 1.2.1 Spectral/hp element Method

Spectral methods and finite element methods (For detailed information on finite element, refer Appendix) are both numerical methods for solving partial differential equations (PDEs). Spectral methods approximate the solution as a series of trigonometric functions, while finite element methods approximate the solution as a series of polynomials.

The spectral/hp element method is a high-order finite element method that combines the geometric flexibility of finite elements with the high accuracy of spectral methods[14]. Spectral methods are typically more accurate than finite element methods, but they are also more computationally expensive. Spectral methods are well-suited for problems with smooth solutions, but they can be inaccurate for problems with sharp features or discontinuities. Finite element methods are less accurate than spectral methods, but they are also less computationally expensive. Finite element methods are well-suited for problems with complex geometries or discontinuities.

Spectral element methods combine the advantages of spectral methods and finite element methods. Spectral element methods use trigonometric functions to approximate the solution, but they do so on a mesh of elements. This allows spectral element methods to achieve high accuracy while also being computationally efficient. In the spectral element method, a polynomial expansion of order p is applied to every elemental domain of a coarse finite element type mesh. This allows the spectral element method to achieve high accuracy even on complex geometries.

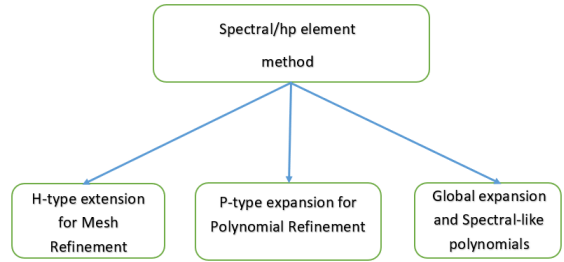

Figure 1.1: Characteristics of Spectral/hp method

The H-type extension is a mesh refinement method that is well-suited for geometrical flexibility. The P-type extension is a polynomial refinement method that is well-suited for high-order approximation. It works by increasing the degree of the polynomials used to represent the solution, which helps to improve the accuracy of the solution over a wider range of spatial scales. The Spectral-like polynomial is a type of polynomial that has ability to achieve the low numerical error.

This section provides a concise and simplified overview of the 1D formulation of the spectral/hp element method [14]

*1. Method of Weighted Residuals*

The method of weighted residuals is a technique for finding approximate solutions to differential equations. It works by first assuming a trial solution of the form of a linear combination of basis functions. The residual is then weighted by a function and integrated over the domain of the differential equation. The coefficients of the trial solution are then chosen to minimize the weighted residual[7]. Consider a general differential equation represented by the linear differential operator L(u) and u (x,t) is the solution to the differential equation [22]

$$L(u) = 0 \qquad (1.1)$$

Given the exact solution u(x,t) to the differential equation, it is possible to define an approximate the numerical solution via trial functions and coefficients as follows [22]

$$u^\delta(x,t) = x_0(x,t) + \sum_{i=1}^{N} \hat{u}_i(t)\phi_i(x) \dots\dots\dots\dots(1.2)$$

However, substituting the approximate solution into the original differential equation (1) does not yield the exact solution [22]

$$L(u^\delta) = R(u^\delta) \tag{1.3}$$

The residual is the difference between the exact solution and the approximate solution [22]

$$R(u^\delta) = u(x,t) - u^\delta(x,t) \tag{1.4}$$

The residual is zero only when the approximate solution is equal to the exact solution. This is a desired property for the numerical solution to have. Therefore, it is important to find a way to enforce this property. This can be done by using the inner product and the weight functions [22]

$$\int_\Omega v_j(x) R \, dx = 0 \tag{1.5}$$

The set of orthogonal weight functions makes the weighted residual over the domain zero. This means that the approximate solution satisfies the original differential equation as the number of degrees of freedom, N, approaches infinity. Therefore, the level of accuracy of the numerical solution is determined by the number of degrees of freedom used in the approximate solution [22]

## 2. Galerkin Formation

The Galerkin method represents an approach rooted in weighted residuals, wherein the trial functions are selected to match the weight functions. This method leverages the integral or weak form of the differential equation to provide an approximation for the solution. Consider the 1D Poisson equation over the domain $\Omega \in [0,1]$ [22]

$$\frac{\partial^2 u}{\partial x^2} + f = 0 \tag{1.6}$$

In its weak form, the above equation can be expressed as follows [22]

$$\int_0^1 v(x) \left( \frac{\partial^2 u}{\partial x^2} + f \right) dx = 0 \tag{1.7}$$

The weak form can be integrated by parts to obtain the following equation [22]

$$\int_0^1 \frac{\partial v}{\partial x}\frac{\partial u}{\partial x}\,dx = \int_0^1 vf\,dx + \left[v\frac{\partial u}{\partial x}\right]_0^1 \tag{1.8}$$

Since the test function v is zero on the Dirichlet boundary x=0, the Neumann boundary condition can be directly implemented. By expressing both the exact solution and the test function as an approximate sum of trial functions (Equation 1.9), it is possible to arrive at the following equation (Equation 1.10) [22]

$$u^\delta = \sum_{i=1}^N u_i(t)\phi_i(x), \quad u^\delta = \sum_{i=1}^N v_i(t)\phi_i(x) \tag{1.9}$$

$$\int_0^1 \frac{\partial v^\delta}{\partial x}\frac{\partial u^\delta}{\partial x}\,dx = \int_0^1 v^\delta f^\delta\,dx + v^\delta(1)g_N \tag{1.10}$$

Finally, to enforce the Dirichlet boundary condition, the approximate solution $u^\delta$ is decomposed into a homogeneous part $u^H$ and a function $u^D$ that satisfies the Dirichlet conditions. This means that $u^H(\partial\Omega) = 0$ and $u^D(\partial\Omega) = g_D$. Substituting this into equation (8), we get the following equation: [22]

$$\int_0^1 \frac{\partial v^\delta}{\partial x}\frac{\partial u^H}{\partial x}\,dx = \int_0^1 v^\delta f^\delta\,dx + v^\delta(1)g_N - \int_0^1 \frac{\partial v^\delta}{\partial x}\frac{\partial u^D}{\partial x}\,dx \tag{1.11}$$

The above equation can be expressed as a system of algebraic equations.

### 3. H-Type Extension

The h-type extension is the feature that gives the spectral/hp element method its geometrical flexibility. This is because the h-type extension allows the mesh to be refined or coarsened in different regions of the domain, depending on the local solution behavior.

The h-type expansion decomposes the solution domain into a set of non-overlapping elements or subdomains, in which local operations can be performed. The size of each element is denoted by h.

$$\Omega = \bigcup_{e=1}^N \Omega^e \tag{1.12}$$

It is often difficult to find a direct expansion in terms of the global bases, also known as global modes, when complex geometries are used. This is because the global modes are not tailored to the specific geometry of the problem. The h-type extension addresses this issue by introducing a standard element, $\Omega_{st}$, that serves as a bridge between the global modes and the local elements. The standard element is defined in terms of a local coordinate, $\xi$, and has local modes, $\varphi_i$. A parametric mapping (Refer Appendix) can then be used to relate the local coordinate to the global coordinate x and vice versa [22]

$$\xi = (X^e)^{-1}(x) \quad x = X^e(\xi) \tag{1.13}$$

Using this mapping, it becomes feasible to articulate the global modes in relation to the local modes.

$$\Phi_i = \left(\varphi_j\left([X_e]^{-1}(x)\right)\right) \ for \ x \in \Omega^e, and \ldots, x \in \Omega^{e+1} \tag{1.14}$$

*4. P-Type Expansion*

The selection of a high-order p-type expansion is the key factor responsible for the exceptional accuracy and fidelity of the spectral/hp element method. Traditional spectral methods employ high-order approximations throughout the entire solution domain through the utilization of Chebyshev, Fourier, or Legendre polynomial expansions [22]

To evaluate the computational efficiency associated with the utilization of a particular polynomial set, we introduce the notion of a mass matrix. This mass matrix emerges from the Galerkin projection of a continuous function f(x) onto the standard region [22]

$$M\hat{u} = f \tag{1.15}$$

*5. Global Assembly*

The final step of the spectral/hp element method is to relate the local modes to the global modes, and is done through global assembly process. Within the Galerkin formulation, computations take place within the elemental standard region, which are subsequently combined to construct the global modes. The transformation from local modes to global modes is performed using the assembly matrix A [22]

$$C = A\hat{u}_g \tag{1.16}$$

Here, $\hat{u}_g$ represents the vector of $N_{dof} - 1$ global coefficients, and $\hat{u}_l$ is a vector consisting of N sets, each containing local coefficients.

### 1.2.2 Spectral/hp solver procedure

Computationally solving a Partial Differential Equation (PDE) across a domain adheres to a specific structure. For the sake of simplicity, let us delve into this process in one dimension. This process entails five primary steps:

*1. Domain Partitioning*

This step divides the domain $\omega$ into a set of non-overlapping local elements $\omega_e$, with a total of $N_e$ elements. The local elements span the entire domain. In one dimension, this can be expressed mathematically as [28]

$$\Omega = \bigcup_{e=0}^{N_e-1} \Omega^e, where \quad \bigcap_{e=0}^{N_e-1} \Omega^e = \emptyset \tag{1.17}$$

## 2. Mapping

Each local element $\Omega_e$ is mapped with a standard element $\Omega_{st}$.

$$\Omega_e = \{x \mid x_e < x < x_{e+1}\} \tag{1.18}$$

$$\Omega_{st} = \{\xi \mid -1 < \xi < 1\} \tag{1.19}$$

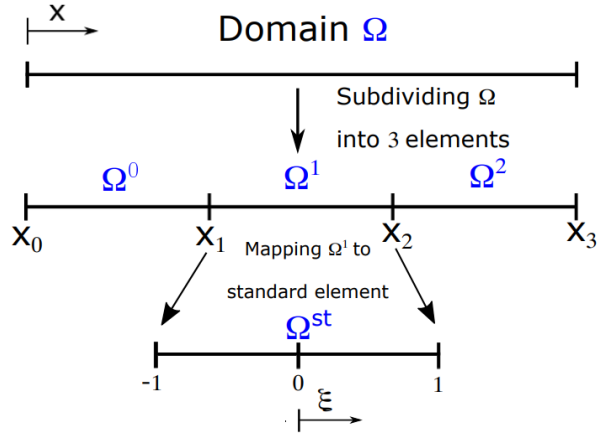

Figure 1.2: Characteristics of Spectral/hp method

The transformation is accomplished using the linear map.

$$x = \chi^e(\xi) = \frac{1-\xi}{2}x_e + \frac{1+\xi}{2}x_{e+1} \tag{1.20}$$

## 3. Expansion of the dependent variables

The numerical solution $u^\delta(x)$ of the partial differential equation is as follows:

$$u^\delta(x) = \sum_{i=0}^{N-1} \hat{u}_j \Phi_j(x) \tag{1.21}$$

where $\hat{u}_j$ is the constant coefficient and $\Phi_j(x)$ are global expansion bases.

In order to derive global expansion bases with non-zero values within a specific local element $\Omega_e$, the initial step involves establishing a set of expansions within the standard element $\Omega_{st}$. The global mode $\Phi_j(x)$ is as follows [28]

$$\Phi_j(x) = \phi_p\left([\chi^e]^{-1}(x)\right) = \phi_p(\xi(x)) \tag{1.22}$$

where, $\phi_p(\xi)$ is the standard expansion base

Note: Lagrange polynomials via Gauss-Lobatto-Legendre quadrature points alternatively serve as the definition for $\phi_p(\xi)$ [30]

$$\phi_p(\xi) = \begin{pmatrix} \frac{1-\xi}{2}(case: p=0) \\ \frac{1-\xi}{2})(\frac{1+\xi}{2})P^{1,1}_{p-1}(\xi)(case: 0>p<P) \\ \frac{1+\xi}{2}(case: p=P) \end{pmatrix} \quad (1.23)$$

Expansion bases is categorized into two different expansions:

*i. Nodal Expansion*

The nodal Lagrange polynomial expansion involves a complete mass matrix, which results in computationally expensive operations.

It also includes N +1 polynomials of the same order N, and Lagrange polynomial expansion falls under this condition, where $\xi_q$ is a set of N + 1 nodes of the Lagrange polynomial.

$$h_p(\xi) = \frac{\sum_{q=0,q\neq p}^{N}(\xi - \xi_q)}{\sum_{q=0,q\neq p}^{N}(\xi_p - \xi_q)}, p = 0,\ldots,N \quad (1.24)$$

*ii. Modal Expansion*

The modal Legendre polynomials are very efficient choice of expansion basis because the mass matrix is diagonal and also has a very good condition number. An expansion set of order N includes as a subset the expansion set of order N-1

*The final steps involves the solution of the matrix system (Equation 15) and global solution assembly (Equation 16)*

Nektar++ is a complex software framework developed the scientific researchers that implements the spectral/hp element methods (Refer Fig. 1.3)

### 1.2.3 Co-kriging

Co-kriging [25] also known as multi-fidelity kriging, uses multiple data sets of varying fidelity to construct a surrogate. The surrogate model incorporates data sets based on their fidelity levels, giving greater importance to more accurate data. This prioritization enhances the predictive accuracy of co-kriging over traditional kriging, particularly when dealing with noisy or incomplete data, enabling better predictions at unobserved locations.

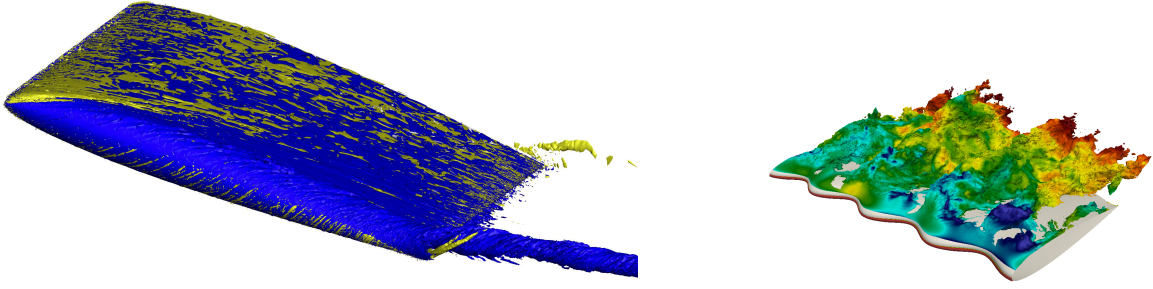

Figure 1.3: (a) Flow simulation for NACA0012 wing tip at Rec = 1.2M using Nektar++ Incompressible flow solver (left) [26] and and (b) DNS Simulation for NACA0012 aerofoil (right) [26]

The method for constructing a co-kriging surrogate presented here is taken from [9], [8], [15]. Co-kriging provides a way to incorporate computationally cheap data into the construction of a kriging surrogate for computationally intensive data. The cheap and intensive data are concatenated to form augmented vectors $\xi$ and $\boldsymbol{f}$ as follows, where $c$ denotes cheap data and $e$ denotes intensive data:

$$\xi = \begin{pmatrix} \xi_c \\ \xi_e \end{pmatrix} \tag{1.25}$$

and

$$f(\xi) = \begin{pmatrix} fc(\xi_c) \\ fe(\xi_e) \end{pmatrix} \tag{1.26}$$

In the context of co-kriging, we approximate the costly data using the less expensive data, scaled by a factor $\rho$, along with the incorporation of a Gaussian process denoted as $Z_d(\xi)$ (d is defined as the difference between the intensive and cheap data). The local mode of the expensive data is defined as follows [15]:

$$Z_e(\xi) = \rho Z_c(\xi) + Z_d(\xi) \tag{1.27}$$

The covariance matrix in co-kriging is defined as follows [15]:

$$C = \begin{pmatrix} cov(fc(\xi_c), fc(\xi_c)) & cov(fc(\xi_c), fe(\xi_e)) \\ cov(fe(\xi_e), fc(\xi_c)) & cov(fe(\xi_e), fe(\xi_e)) \end{pmatrix} \tag{1.28}$$

with matrix elements defined as follows:

$$cov(fc(\xi_c), fc(\xi_c)) = \begin{pmatrix} cov(Z_c(\xi_c), Z_c(\xi_c)) \\ \sigma_c^2 R_c(\xi_c, \xi_c) \end{pmatrix} \tag{1.29}$$

$$\text{cov}(fe(\xi_e), fc(\xi_c)) = \begin{pmatrix} \text{cov}(\rho Z_c(\xi_c) + Z_d(\xi_c), Z_c(\xi_c)) \\ \rho \sigma_c^2 R_c(\xi_c, \xi_e) \end{pmatrix} \quad (1.30)$$

$$\text{cov}(fe(\xi_e), fe(\xi_e)) = \begin{pmatrix} \text{cov}(\rho Z_c(\xi_e) + Z_d(\xi_e), \rho Z_c(\xi_e)) + Z_e(\xi_e) \\ \rho^2 \text{cov} Z_c(\xi_e), Z_c(\xi_e) + \text{cov} Z_d(\xi_e), Z_d(\xi_e) \\ \rho^2 \sigma_c^2 R_c(\xi_e, \xi_e) + \sigma_d^2 R_d(\xi_e, \xi_e) \end{pmatrix} \quad (1.31)$$

The fundamental process for constructing the co-kriging surrogate closely mirrors that of ordinary kriging. Initially, the hyperparameters for the inexpensive data are determined, and subsequently, the parameters for the costly data are determined by establishing a connection between the inexpensive data and the costly data through a vector $d$ which is defined as follows [15]

$$d = \rho f_e - \rho f_c(\xi_e) \quad (1.32)$$

The notation $f_c(\xi_e)$ represents the points located within the $\xi_e$ positions identified in the inexpensive data set.

In order to determine the hyperparameters for '$d$' and the scaling parameter '$\rho$', we utilize the natural logarithm of the likelihood for '$d$', which is expressed as follows [15]

$$Ln(\mu_d, \sigma_d^2, \Theta) = -\frac{Ne}{2}\ln(2\pi) \; - \; \frac{Ne}{2}\ln(\sigma_d^2) \; - \; \frac{1}{2}\ln|R_d(\xi_e, \xi_e)| \; - \; \frac{(d - 1\mu_d)^T R_d(\xi_e, \xi_e)^{-1}(d - 1\mu_d)}{2\sigma^2} \quad (1.33)$$

Closed-form solutions for $\mu_d$ and $\sigma_d^2$ can be obtained by maximizing equation 1.33, leading to the resultant outcomes [15]

$$\mu_d = \frac{1^T R_d(\xi_e, \xi_e)^{-1} d}{1^T R_d(\xi_e, \xi_e)^{-1} 1} \quad (1.34)$$

$$\sigma_d^2 = \frac{(d - 1\hat{\mu}_d)^T R_d(\xi_e, \xi_e)^{-1}(d - 1\hat{\mu}_d)}{Ne} \quad (1.35)$$

By replacing equations 1.34 and 1.35 into equation 1.33, we derive the concentrated likelihood equation for $d$, as follows [15]

$$Ln(\Theta) = -\frac{Ne}{2}\ln(2\pi) \; - \; \frac{Ne}{2}\ln(\hat{\sigma}_d^2) \; - \; \frac{1}{2}\ln|R_d(\xi_e, \xi_e)| \quad (1.36)$$

The co-kriging predictor is obtained by utilizing the augmented natural logarithm likelihood equation [15]

$$fe(\xi) = \hat{\mu} + c^T \left(C^{-1}\right)(f - 1\hat{\mu}) \tag{1.37}$$

$$C = \begin{pmatrix} \hat{\rho}\hat{\sigma}_c^2 k(\xi_c, \xi^i) \\ \hat{\rho}^2 \hat{\sigma}_c^2 k(\xi_e, \xi^i) + \hat{\sigma}_d^2 k(\xi_e, \xi^i) \end{pmatrix} \tag{1.38}$$

Minimizing the mean square error is the method employed to determine the model's uncertainty, leading to the resulting estimation of uncertainty [15]

$$s^2(\xi) = \rho^2 \hat{\sigma}_c^2 + \hat{\sigma}_d^2 - c^T \left(C^{-1}\right) c + \frac{(1 - 1^T C^{-1} c)^2}{c^T \left(C^{-1}\right) c} \tag{1.39}$$

### 1.2.4 Deep Neural Networks

Deep Neural Network (DNN) [23] is a type of artificial neural network designed to model and solve complex tasks. DNN is composed of multiple interconnected layers of artificial neurons, and these layers are organized into three sections, Input layer, Hidden layers and Output layer.

Input Layer receives the raw input data, and Output Layer produces the outputs/predictions.

Hidden Layers: These layers are positioned between the input layer and the output layer performs complex transformations and feature extraction.

*Activation Function*

To introduce non-linear characteristics into the neural network, every neuron undergoes a transformation by passing its initial output through an activation function to obtain the final output. This activation function is versatile and can be chosen from a range of arbitrary functions. The choice of activation function affects the model training and performance. This approach expedites the training process, and additional complexity can be introduced by incorporating extra neurons and layers as needed.

The below mentioned function provides an overview of various activation functions (Refer Appendix):

Linear function: F(x) = x & Hyperbolic function F(x) = $\tanh(x)$

Rectified Linear Unit (ReLU) function:

$$F = \begin{pmatrix} x, & x \geq 0 \\ 0, & x \leq 0 \end{pmatrix} \tag{1.40}$$

*Hyperparameter optimisation*

Hyperparameters are settings that control the training process and the architecture of a neural network. They are set before the training process begins. Optimizing hyperparameters is essential for achieving better performance in deep neural networks, and by finding the optimal set of hyperparameters, one can improve the accuracy, speed, and stability of the model (Refer Appendix).

*Deep Neural Network Training*

After creating the neural network, the next step is to train it using the dataset with known outputs. The training process involves adjusting and converging the weights of each neuron. To enhance the model, the initial step is to evaluate its performance relative to the known solution. This assessment is carried out using a "Loss function." While loss functions can be customized based on the specific situation, there are some commonly used ones such as Mean Squared Error for modeling applications.

Once the model performance has been quantified, the next step involves optimizing the weights to enhance future predictions, and is typically achieved through a process called "fine-tuning". A neural network is trained by passing the training data through the network multiple times. Each pass through the data is called an epoch, and in each epoch, the network weights are adjusted to improve the fit to the data.

*Multi-Fidelity Deep Neural Network*

Deep learning is being employed across various domains, including modeling physical, communication, and biological systems [16]. The algorithms are known for their significant demand for high-quality training data, which poses challenges for practical applications due to the high cost of generating such data [16]. In engineering, producing sufficient high-fidelity data is often not feasible, leading to compromised prediction accuracy [16]. This issue is particularly critical in high-dimensional uncertainty quantification, where accurately quantifying input uncertainty effects on system outputs requires a prohibitive number of high-fidelity samples [16].

In order to reduce the dependency on costly HF data, multi-fidelity techniques are particularly used in surrogate modeling and Uncertainty Quantification [16]. This method leverage both high and low-fidelity data, utilizing the readily available and inexpensive low-fidelity data for better model prediction accuracy [16]. Various multi-fidelity approaches, including multi-fidelity response surface models, multi-fidelity artificial neural networks, and multi-fidelity Gaussian processes, have been explored [16]. However, challenges still remain ineffectively bridging the gap between low- and high-fidelity data, as well as in dealing with sparse data and high-dimensional problems [16].

Recent advancements include the development of Multi-fidelity neural networks for surrogate modeling and Uncertainty Quantification [16]. The model can adapt to both low and high dimensions and is being applied in various problems, such as estimating high-fidelity coefficients, solving Bayesian inverse problems, and addressing the time-dependent issues [16]. Also, some approach do involve constructing networks that directly incorporate corrections between low- and high-fidelity data, aiming to enhance the prediction accuracy [16].

The architecture for multi-fidelity deep neural networks explained in this research work simplifies the previous designs by integrating a single correction sub-network [16]. It eliminates the need for separate assumptions for linear and nonlinear corrections, aligns with Multi-fidelity methodology's mathematical principles, and simplifies the training and testing process, especially for high-dimensional challenges [16]. This is the first attempt at addressing high-dimensional aleatory UQ problems with an MF-DNN [16].

*Why Multi-Fidelity Deep Neural Network??*

To reduce the dependence on the high-fidelity numerical simulations that are computationally very expensive for the purpose of uncertainty quantification, multifidelity approaches have become prominent because of their capacity to merge data of varying fidelity [17]. The idea of multi-fidelity methods is to ensure the precision of surrogate models by merging abundant low-fidelity data with a scarce quantity of High-fidelity data [17]. Earlier research did introduced multi-fidelity polynomial chaos expansion and multi-fidelity Monte Carlo techniques, these approaches often struggle to effectively bridge the gap between low-fidelity and High-fidelity data, especially when dealing with high-dimensional datasets [17]. The multi-fidelity deep neural networks has recently been recognized as a promising solution for surrogate modeling and UQ challenges because of the deep neural network natural capacity for universal function approximation [17].

Motamed did developed two distinct neural networks using the bi-level training data to assess the variability in two-dimensional function outputs, considering input uncertainties [21], [17]. Meng and Karniadakis [19], [17] 4did introduced composite neural network framework featuring three sub networks: one for LF predictions, another for HF predictions assuming linear relationships between low-fidelity and high-fidelity data, and a third for high-fidelity predictions under nonlinear assumptions. This composite model was tested against several benchmarks [19], [17]. Following this, Zhang et al. [32], [17] applied the multi-fidelity deep neural network architecture from Meng and Karniadakis's research for the aerodynamic optimization of the RAE2822 airfoil and the DLR-F4 wing-body configuration.

Based on the literature review, it is evident, this is the first application of multi-fidelity deep neural network in solving the UQ problem for the turbomachinery flows [17].

## 1.3 Problem Statement

High-fidelity simulation data are very expensive and requires HPC clusters to perform the simulation. It requires large GPU hours to run these high-fidelity simulations. For data predictions, we do use Kriging and Co-kriging. Co-kriging demands more computational resources than kriging because it requires inverting a matrix of size (Nc + Ne) by (Nc + Ne), compared to the smaller Ne by Ne matrix inversion needed for kriging, where Ne represents the count of high-cost samples and Nc denotes the count of low-cost samples [15]. Additionally, co-kriging necessitates a comprehensive search to determine the scaling parameter, $\rho$. Therefore, the implementation of co-kriging is justified only when the computational expenses associated with running the CFD code surpass those involved in setting up the kriging model [15].

Kerry et al. [15] have perfomed the Kriging and Co-Kriging with Adaptive Sampling to solve the problem of Uncertainty Quantification in Computational Fluid dynamics, and have showed the advantage of combining low and high fidelity CFD simulations to reconstruct an accurate meta-model. The cfd simulation were carried out using STAR-CCM+. In our research work, as already discussed, Co-kriging technique has been used to obtain the precise data predictions for the lift and drag within the confined domain, and the expensive data has been obtained using Nektar++ (Spectral element package) and cheap data has been obtained using the XFOIL package.

To further minimize the reliance on high-fidelity numerical simulations for Uncertainty Quantification, a multi-fidelity strategy has been adopted [17]. The core principle behind the multi-fidelity method is to maintain the accuracy of the surrogate model predictions by leveraging large number of low-fidelity data complemented by a smaller set of high-fidelity data [17]. The Multi-fidelity deep neural network is particularly chosen for its potential in addressing surrogate modeling and Uncertainty quantification challenges, owing to the deep neural network inherent ability to approximate functions universally [17]. Additionally, error analysis has been conducted to assess the performance of both Co-kriging and multi-fidelity deep neural network, revealing that the multi-fidelity deep neural network method has outperformed Co-kriging in effectiveness [17], [16].

The Multi-dimensionality test has been performed using MF-DNN to evaluate the performance of the multi-fidelity deep neural network in solving complex UQ problems.

The research creates the methodology for quantifying uncertainty in computational fluid dynamics with a focus to minimize the required number of samples. This significantly lowers the computational costs [15]. The research also shows the the benefits of integrating low and high fidelity CFD simulations by developing precise meta-models [15]. The uncertainty quantification framework used in the research is essential for the design and optimization of advanced turbomachinery considering high-dimensional input uncertainties [17]. Based on the literature review, it is also evident, the application of multi-fidelity deep neural network to solve the UQ problem in turbomachinery flows is new [17].

# Chapter 2

# Methodology

## 2.1 Overview of Nektar++: High-Fidelity Simulation

The spectral/hp method exhibits exceptional convergence characteristics, rendering it highly appealing for applications in fluid dynamics. Nektar++ was formulated to enhance accessibility to scientists and engineers, both in academic and industrial settings. Nektar++ is a software framework that not only integrates the intricate mathematical methodology of the approach into an open-source C++ framework but also furnishes pre-existing PDE solvers. These solvers can be utilized in their original form for developing solutions designed to address scientific and engineering challenges. The reason behind conducting the high-fidelity simulations is to obtain the expensive data [15] for Uncertainty quantification.

The Nektar++ codebase is depicted in Figure 2.1, and is founded on the description of the spectral/hp method described in the below mentioned equation [24]. The Nektar++ library encompasses a total of six core embedded libraries, as illustrated in Figure 2.2 [13]

$$u(\mathbf{x}) = \sum_{e \in \mathcal{E}} \overbrace{\underbrace{\sum_{n \in \mathcal{N}} \phi_n^e(\mathbf{x}) \hat{u}_n^e}_{\text{LocalRegions library}}}^{\text{MultiRegions library}} = \sum_{e \in \mathcal{E}} \sum_{n \in \mathcal{N}} \phi_n^{std} \underbrace{\overbrace{([\mathcal{X}^e]^{-1}(\mathbf{x}))}^{\text{SpatialDomains library}}}_{\text{StdRegions library}} \hat{u}_n^e$$

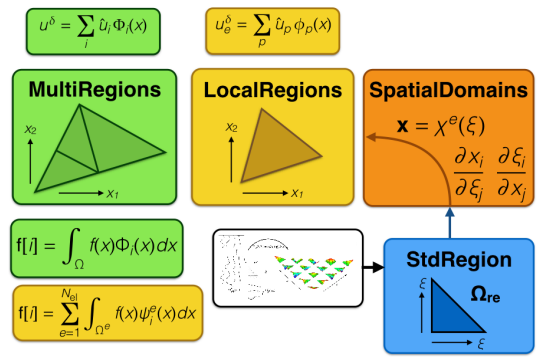

Figure 2.1: Nektar++ codebase is structured into three main components: solvers, libraries, and utilities [24]

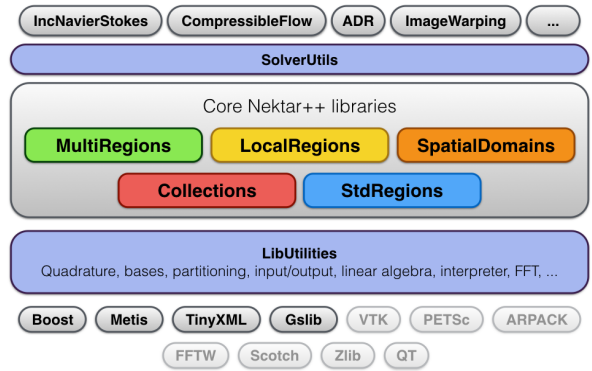

Figure 2.2: Main libraries of Nektar++ codebase [4]

### 2.1.1 SolverUtils Library

The SolverUtils library is responsible for developing high-level components that are intended to be used across various solvers. These components encompass time-stepping algorithms, standardized modules for incorporating diffusion and advection terms, as well as input/output (I/O) capabilities [6].

### 2.1.2 MultiRegions Library

The MultiRegions consolidate the solution across the entirety of the computational domain. GlobalLinSys class provides a suite of efficient algorithms tailored to solve linear system equations, encompassing methodologies like static condensation. In parallel, the ExpList class is responsible for generating the expansion list pertaining to physical variables, $u,v,p$ [6].

### 2.1.3 LocalRegions Library

The LocalRegions module provides essential operations that are carried out within the physical computational domain.

### 2.1.4 SpatialDomains Library

The SpatialDomains library is employed for the global assembly process. Within this library, **Geometry classes** are employed to represent the geometry of individual elements within the physical space. **MeshGraph** class take on the responsibility of constructing the solution domain $\Omega$ and the individual elements $\Omega_e$ based on the input file, and **BoundaryCondition** class manages and enforces the constraints imposed by a given boundary condition on a particular element [6].

### 2.1.5 StdRegions Library

The StdRegions library provides fundamental operations such as integration and differentiation for a variety of geometries, including 1D segments, 2D triangles, unit squares, and 3D tetrahedra, hexahedra, prisms, and pyramids. These standard regions are currently implemented in the Nektar++ framework.

### 2.1.6 LibUtilities Library

The LibUtilities library provides a diverse set of functionalities, establishing the fundamental building blocks essential for the implementation of spectral/hp element methods. These components include polynomial bases defined on standard regions ($\phi_i$), the utilization of coordinate distributions ($\xi_j$) in numerical integral and differential formulations, the derivative matrices associated with the basis functions, and the incorporation of the quadrature weight matrix [6].

### 2.1.7 Preprocessing

NekMesh serves the purpose of converting a provided mesh into the Nektar++ proprietary file format. NekMesh is also utilized to create high-order meshes within a domain specified by a specific file format.

### 2.1.8 Postprocessing

Nektar++ features an application known as FieldConvert, which empowers users to convert the output files generated by a Nektar++ solver into various file formats. These converted files can then be visualized using diverse visualization tools like Paraview or VisIt. Additionally, FieldConvert offers the capability to perform various manipulations and computations on the data files produced by any of the Nektar++ solvers.

The process of utilizing Nektar++ can be visualized through the flowchart (Fig. 2.3).

### 2.1.9 Numerical Setup

The analysis of flow around the NACA0012 airfoil and the acquisition of data for the coefficient of lift and drag were carried out utilizing an incompressible Navier-Stokes solver, named IncNavierStokesSolver. To execute the solver successfully, it is essential to have the mesh and session files in .xml format, which contain crucial information regarding the solver settings and boundary conditions.

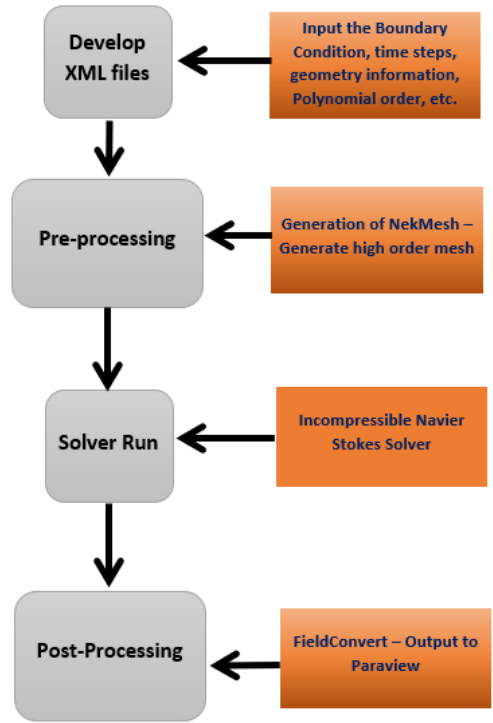

Figure 2.3: Simulation procedure for Nektar++

### 2.1.10 Mesh Generation

Initially, the mesh was generated using Gmsh software. Subsequently, it underwent conversion into the .xml format through the utilization of the NekMesh utility. This conversion process ensured that the mesh became compatible with the Nektar++ solver, enabling the comprehensive analysis of flow characteristics for various polynomial orders (For further information, please refer Appendix).

### 2.1.11 Session file

The session file (in .xml format) contains information about the Expansions, Solver setup, Simulation parameters, Variables, Boundary, and initial conditions required to perform the 2-Dimensional simulations (Refer Appendix) in Nektar++ (For further information on session file, please refer Appendix).

### 2.1.12 Coefficient of Lift and Drag

The Coefficient of lift ($C_L = \frac{L}{\frac{1}{2}\rho V^2 A}$) represents the lift generated by the airfoil relative to the fluid density around it, whereas, the Coefficient of drag quantifies the drag or resistance of the airfoil in the fluid environment ($C_D = \frac{D}{\frac{1}{2}\rho V^2 A}$). The lift and drag coefficient is represented by ($C_L$) and ($C_D$) term respectively.

## 2.2 Overview of XFOIL: Low-Fidelity Simulation

XFOIL is a panel method that uses a simplified approach to solve the equations of motion for an airfoil. This makes it a faster and easier-to-use tool than other methods, such as the finite element method. However, it is also less accurate.

The XFOIL 6.99 package was used to evaluate the inviscid flow around the NACA0012 airfoil at an angle of incidence of $0.5° \leq \alpha \leq 7.5°$, ($\alpha$ being the angle of incidence), with an increment of $\delta\alpha = 0.5°$ and Reynolds number of Re=$10^5$.

The inviscid solver evaluates the Pressure coefficient (Cp) plot, Coefficient of lift and Coefficient of drag coefficients. The lift and drag coefficient values at incidence angle of 1°, 3°, 5° and 7° has been used to perform the Co-kriging data fusion and adaptive sampling for Uncertainty quantification.

The below mentioned table consists the Coefficient of lift and drag data obtained from the XFOIL simulation.

Table 2.1: Dataset for the Coefficient of lift & Coefficient of drag obtained from XFOIL simulation

| Angle of Attack | Coefficient of Lift | Coefficient of Drag |
|---|---|---|
| 1° | 0.4797 | 0.01937 |
| 3° | 0.6368 | 0.01923 |
| 5° | 0.8192 | 0.01891 |
| 7° | 0.9572 | 0.01965 |

## 2.3 Dataset for Uncertainty Quantification using Co-Kriging

Comparison tables for the coefficient of lift and drag values has been constructed using the data gathered from both high-fidelity (Nektar++ simulation) and low-fidelity simulations (XFOIL simulation). The dataset has been utilized to generate the Co-kriging data fusion and adaptive sampling plot.

Table 2.2: Coefficient of lift values at different angle of attack from High-fidelity & Low-fidelity simulation

| Angle of Attack | Coefficient of Lift (XFOIL) | Coefficient of Lift (Nektar++) |
|---|---|---|
| 1° | 0.4797 | 0.6270 |
| 3° | 0.6368 | 0.7623 |
| 5° | 0.8192 | 0.9333 |
| 7° | 0.9572 | 0.9929 |

Table 2.3: Coefficient of drag values at different angle of attack from High-fidelity & Low-fidelity simulation

| Angle of Attack | Coefficient of Drag (XFOIL) | Coefficient of Drag (Nektar++) |
|---|---|---|
| 1° | 0.01937 | 0.01891 |
| 3° | 0.01923 | 0.01854 |
| 5° | 0.01891 | 0.01809 |
| 7° | 0.01965 | 0.01904 |

## 2.4 Co-Kriging Data Fusion

Co-kriging data fusion is a powerful technique deployed within the computational fluid dynamics for the purpose of Uncertainty Quantification. It serves the dual objective of enhancing the precision and dependability of CFD simulations while concurrently quantifying the uncertainties linked to the outcomes. This approach involves combining data from both high-quality and low-fidelity simulations to achieve more accurate data prediction and a better understanding of the uncertainties associated with these simulations.

## 2.5 Co-Kriging Adaptive Sampling

Adaptive sampling for uncertainty quantification [29] is used to efficiently reduce uncertainties in CFD predictions. It accomplishes this by strategically choosing new sampling locations based on the existing state of the model, with the primary aim of minimizing the prediction uncertainty. Within the domain of CFD simulations, uncertainties can originate from the diverse sources, encompassing boundary conditions, initial conditions, model parameters, and numerical approximations. These uncertainties introduce variations in the simulation outcomes, potentially impacting the precision of predictions, particularly in scenarios involving intricate fluid flow dynamics.

Selecting infill criteria or adaptive sampling strategies is crucial for developing a suitable surrogate model [15]. Given the high computational costs involved, it's beneficial to sample the stochastic space efficiently and accurately with the minimum number of new sample points needed to precisely represent the output's behavior [15]. Various methods exist for updating the surrogate, including minimizing the predictor, enhancing the surrogate based on expected improvement, utilizing the uncertainty in the predictor, and employing goal seeking through the Maximum Likelihood Estimation (MLE) [15]. Techniques that focus on minimizing the predictor, aiming for expected improvement, or targeting the MLE are generally applied in optimization issues to identify minima[15]. However, in Uncertainty Quantification (UQ) in Computational Fluid Dynamics (CFD), these methods may not be always the best fit due to the potential for highly non-linear output behavior driven by the fluid dynamics non-linearity, where the output's minimum may not represent the most significant or relevant feature for creating an accurate surrogate (For the detailed mathematics involved in Adaptive sampling, please refer Appendix).

The core idea of adaptive sampling in CFD revolves around the strategic selection of new sampling points within the computational domain [15]. This selection is informed by the current state of the simulation and the overarching goal of minimizing prediction uncertainty. Adaptive sampling methodologies leverage information gleaned from the existing simulation results to pinpoint regions characterized by high prediction uncertainty or substantial variations [15]. Moreover, these techniques are capable of considering the computational expenses associated with conducting additional simulations. This consideration is particularly relevant since CFD simulations often entail significant computational costs.

## 2.6 Surrogate Modelling and Uncertainty Quantification

A multi-fidelity deep neural network model has been implemented that is designed to learn from a limited quantity of high-fidelity data supplemented by a larger quantity of low-fidelity data. It is a challenge in engineering problems due to the high costs associated with generating or obtaining high-fidelity data [16]. Unlike traditional approaches that utilize parallel structures to independently model the linear and non-linear relationships between high-fidelity and low-fidelity data, the below mentioned multi-fidelity Deep Neural network model integrates these tasks within a single sub-network [16]. This integration allows the network to adaptively identify and learn both linear and non-linear correlations without the need for manual intervention in adjusting output weights for separate correction networks [16]. This self-learning capability significantly enhances the network's training efficiency and applicability, managing complex, high-dimensional problems [16].

The widely used comprehensive correction in bridging low-fidelity and high-fidelity data is [16]:

$$\hat{y}_{HF} = \rho(X) \cdot y_{LF}(X) + \delta(X) \tag{2.1}$$

where $\hat{y}_{HF}$ represents the model prediction values on high-fidelity data points, $\rho$ is the multiplicative correction surrogate, $y_{LF}$ represents the label values of low-fidelity data points and $\delta$ represents the additive correction surrogate [16]. The multiplicative correction $\rho$ can be either a constant or non-constant value, which represents the linear or non-linear correction between $\hat{y}_{HF}$ and $y_{LF}$ [16]. The above shown comprehensive correction is shown in the following form: [16]:

$$\hat{y}_{HF} = F(y_{LF}(X), X) \tag{2.2}$$

where F represent both the non-linear and linear correlation between the low-fidelity and high-fidelity data [16]. The idea for the proposed multi-fidelity deep neural network is that it should consist of two sub-networks, one for the low-fidelity deep neural network to approximate the values of $y_{LF}$, and one for the correction deep neural network to predict the $\hat{y}_{HF}$ based on the expression (2.2) [16].

The architecture of the new Multi-Fidelity Deep Neural Network is shown in the below mentioned figure Fig. 2.4 [17], [16]. The framework assumes the presence of a large set of low-fidelity training data, denoted by $y_{LF}(X_{LF})$, where $X_{LF} \in R$, and a comparably smaller set of high-fidelity training data, represented by $y_{HF}(X_{HF})$, with $X_{HF} \subseteq R$ [16]. The cost function is the mean square error (MSE), which computes the discrepancy between the predicted outputs and the true values [16]. The gradient information for this cost function is obtained through automatic differentiation and is important for updating the network parameters[16]. The minimization of the prediction error for the multi-fidelity deep neural network model is achieved through the application of gradient-based optimization algorithms [16]. The ADAM and L-BFGS optimizers are utilized because of their superior generalization performance when compared to other optimization strategies [16]. To prevent overfitting, the $L_2$ regularization loss has been implemented to minimize the overall loss by summing the squared magnitudes of the network weight coefficients [16].

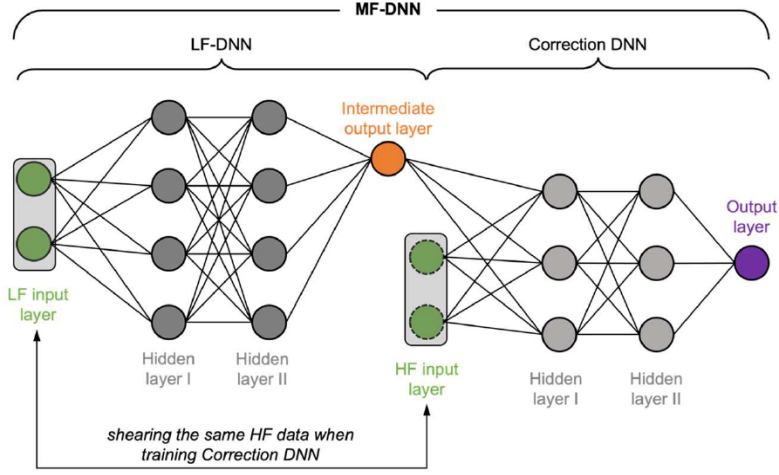

Figure 2.4: Architecture of the Multi-fidelity Deep Neural Network [16]

The low-fidelity deep neural network and the correction deep neural network are trained in sequence and this process is important from the prospective of programming and training efficiency [16]. The specific definition of the loss function for the low-fidelity deep neural network is shown below [16]:

$$L_l = \frac{1}{M} \sum_{i=1}^{M} (\hat{y}_{LF}(X_{LF}; \theta_{NN}) - y_{LF}(X_{LF}))^2 \tag{2.3}$$

$$\theta_{NN} = \{(w_h, b_h)\}_{h=1}^{H+1} \tag{2.4}$$

$L_l$ is the training loss for the low-fidelity deep neural network, $M$ represents the number of low-fidelity deep neural network training, $\theta_{NN}$ are the neural network parameters, $w_h$ denotes the weights of the $h$-th layer, $b_h$ refers to the biases of the $h$-th layer, and $H$ indicates the number of hidden layers [16]. The trained low-fidelity deep neural network is then utilized as a static surrogate model, which implies it remains unaltered and is not subject to retraining during the training phase of the correction deep neural network [16]. The loss function for the correction deep neural network is defined as follows [16]:

$$L_c = L_h + L_r \tag{2.5}$$

$$L_h = \frac{1}{P} \sum_{i=1}^{P} (\hat{y}_{HF}(X_{HF}; \theta_{NN}) - y_{HF}(X_{HF}))^2 \tag{2.6}$$

$$L_r = \lambda \sum w_h^2 \tag{2.7}$$

$L_c$ denotes the total training loss of the correction deep neural network, $L_h$ corresponds to the mean squared error of the correction deep neural network [16]. The term $L_r$ stands for the $L_2$ regularization loss of the correction deep neural network, $P$ represents the number of high-fidelity training data points, and $\lambda$ is the control parameter of the $L_2$ regularization loss [16]. To enable the correction deep neural network with the capability to distinguish the differences between the predictions from the low-fidelity deep neural network and corresponding high fidelity deep neural network, the high-fidelity variables $X_{HF}$ are incorporated during the training phase [16]. On including the high-fidelity variables in the training process, the correction deep neural network obtained the ability to comprehend and capture the variances between the low-fidelity deep neural network predictions and the actual high-fidelity labels, thus facilitating its precision in correcting and refining the predictions accordingly [16].

The network hyperparameters consists of numbers of the hidden layers, number of neurons in hidden layers, learning rate and others were optimized by the Bayesian optimization algorithm [16]. This optimization strategy has been selected because of its reputation as an effective global optimizer for noisy and complex problems [16].

The performance of the refined Multi-Fidelity Deep Neural Network model is then assessed with the K-fold cross-validation strategy [16]. Rectified linear unit (ReLU) activation function was implemented rather than using the conventional hyperbolic tangent activation function to achieve more effective approximation of both linear and non-linear relationships within the single sub-network [16].

The Rectified Linear Unit (ReLU) activation function is mainly used for enhancing the training speed and effectiveness of deep neural networks, especially for datasets with high dimensionality [16]. The construction and training of the multi-fidelity deep neural network model are conducted within the TensorFlow environment [16]. The detailed training steps for the multi-fidelity deep neural network model are as follows [16]:

1. Establishing the initially fully connected low-fidelity deep neural network[16].

2. Fine-tuning the low-fidelity deep neural network hyperparameters using Bayesian optimization method [16].

3. Training the low-fidelity deep neural network on $M$ realizations from the low-fidelity dataset $Q_{LF} = \{Y_{LF}^{(1)}, \ldots, Y_{LF}^{(M)}\}$, using ADAM and L-BFGS optimizer [16].

4. Creating the initial correction deep neural network and calling the trained low-fidelity deep neural network as a sub-module [16].

5. Optimizing the hyperparameters in the correction deep neural network via Bayesian optimization [16].

6. Training the correction deep neural network using the high-fidelity dataset $Q_{HF} = \{v_{HF}^{(1)}, \ldots, v_{HF}^{(N)}\}$ with $M >> N$, using the ADAM and L-BFGS optimizers [16]

7. Validating and testing the accuracy of the trained multi-fidelity deep neural network using the K-fold cross-validation method [16].

# Chapter 3

# Results and Discussions

## 3.1 High-Fidelity Simulations

In the realm of Computational Fluid Dynamics, the fundamental objective of Uncertainty Quantification is to systematically assess and control uncertainties stemming from diverse origins, including fluctuations in initial conditions, boundary conditions, and modeling parameters. One effective strategy for achieving this goal involves the utilization of varying polynomial orders within the Nektar++ framework. These chosen polynomial orders fundamentally dictate the degree of approximation and intricacy within the numerical solution.

An intriguing investigation has been conducted to observe how varying the polynomial order influences the precision and dependability of lift predictions. The analysis involved examining the lift coefficient versus angle of attack plot, which served as a means to evaluate the ability of the CFD simulations to capture fluid flow behavior, particularly at different angles of attack. This is important to obtain the precise data for further research investigation. The comparison plot (Figure 3.2) offers valuable insights into the influence of numerical approximations, represented by the chosen polynomial orders, on the accuracy of lift predictions. This analysis aids in ascertaining whether higher polynomial orders yield more accurate outcomes or if a lower order suffices. This comprehensive examination is pivotal in the realm of uncertainty quantification, ensuring that CFD simulations deliver trustworthy results, even in the presence of inherent uncertainties.

### 3.1.1 Effect of Polynomial distribution on Coefficient of lift

Polynomial distributions of higher order provides more accurate representations of airfoil surfaces. These distributions are mathematical functions that represent the shape of the aerofoil surface. The coefficients of the polynomial are determined by fitting the polynomial to a set of data points, such as computational fluid dynamics simulation results. The coefficient of lift is then calculated from the polynomial distribution.

The higher the order of the polynomial, the more accurate the representation of the aerofoil surface. This is because a higher-order polynomial has more terms, which allows it to better fit the data points. As a result, the coefficient of lift calculated from the higher polynomial distribution is more accurate [11], [31]. The same can been seen from Figure 3.2 and Table 3.1. In addition to that, the lift and drag data obtained from the polynomial order 6 is being used for further research investigation because of its precision as compared to the data obtained from polynomial order 2 and 4.

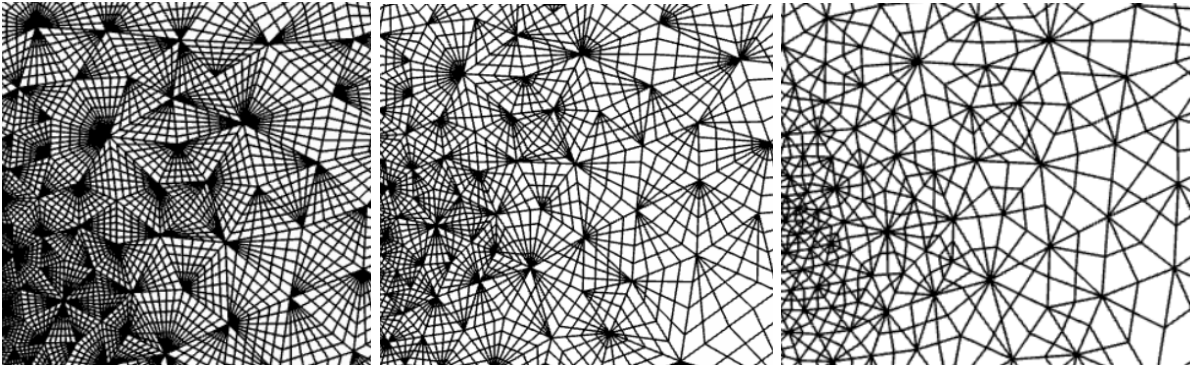

Figure 3.1: Polynomial distribution with same element size for different NumModes (a) NumModes = 7 (Left) (b) NumModes = 5 (Middle) (c) NumModes = 3 (Right)

Table 3.1: Coefficient of lift dataset for different polynomial order

| Parameter | AOA1 | AOA3 | AOA5 | AOA7 |
|---|---|---|---|---|
| HFP2 | 0.571 | 0.7086 | 0.8633 | 0.9666 |
| HFP4 | 0.593 | 0.7313 | 0.9083 | 0.9828 |
| HFP6 | 0.627 | 0.7623 | 0.9333 | 0.9929 |

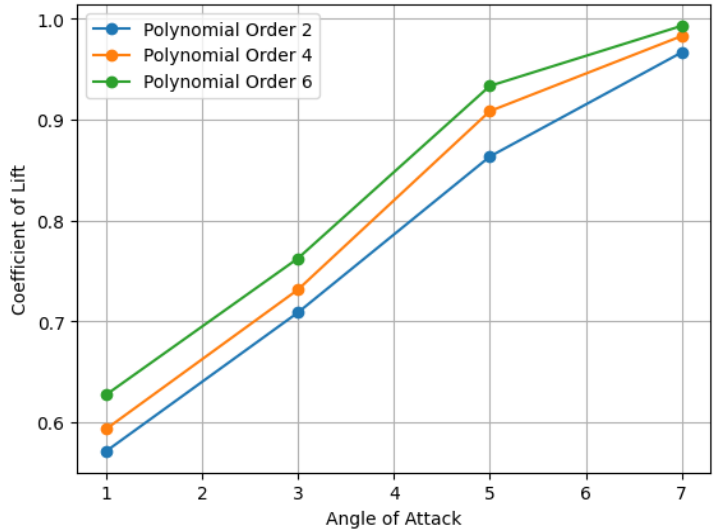

Figure 3.2: Coefficient of Lift vs Angle of attack comparison plot for different polynomial order

The computational fluid dynamics results provide a comprehensive analysis of fluid flow behavior under diverse conditions, considering both angles of attack and the complexity of the numerical models utilized. The choice of angle and polynomial order [1] reflects a thoughtful balance between accuracy and computational resource requirements in CFD research. Specifically, the following CFD findings pertain to angles of attack at 1°, 3°, 5°, and 7°, obtained using polynomial orders of 6, 4, and 2, respectively for 1 million steps (i.e., Number of steps).

Initially, computational fluid dynamics results were acquired for polynomial order 6, covering a range of angles of attack over a span of 0.3 million steps. However, these initial results did not exhibit convergence and revealed discrepancies in the flow visualization. Consequently, the simulations were extended to 1 million steps. Following are the computational fluid dynamics outcomes for different angles of attack, obtained through polynomial orders of 6, encompassing a duration of 0.3 million steps.

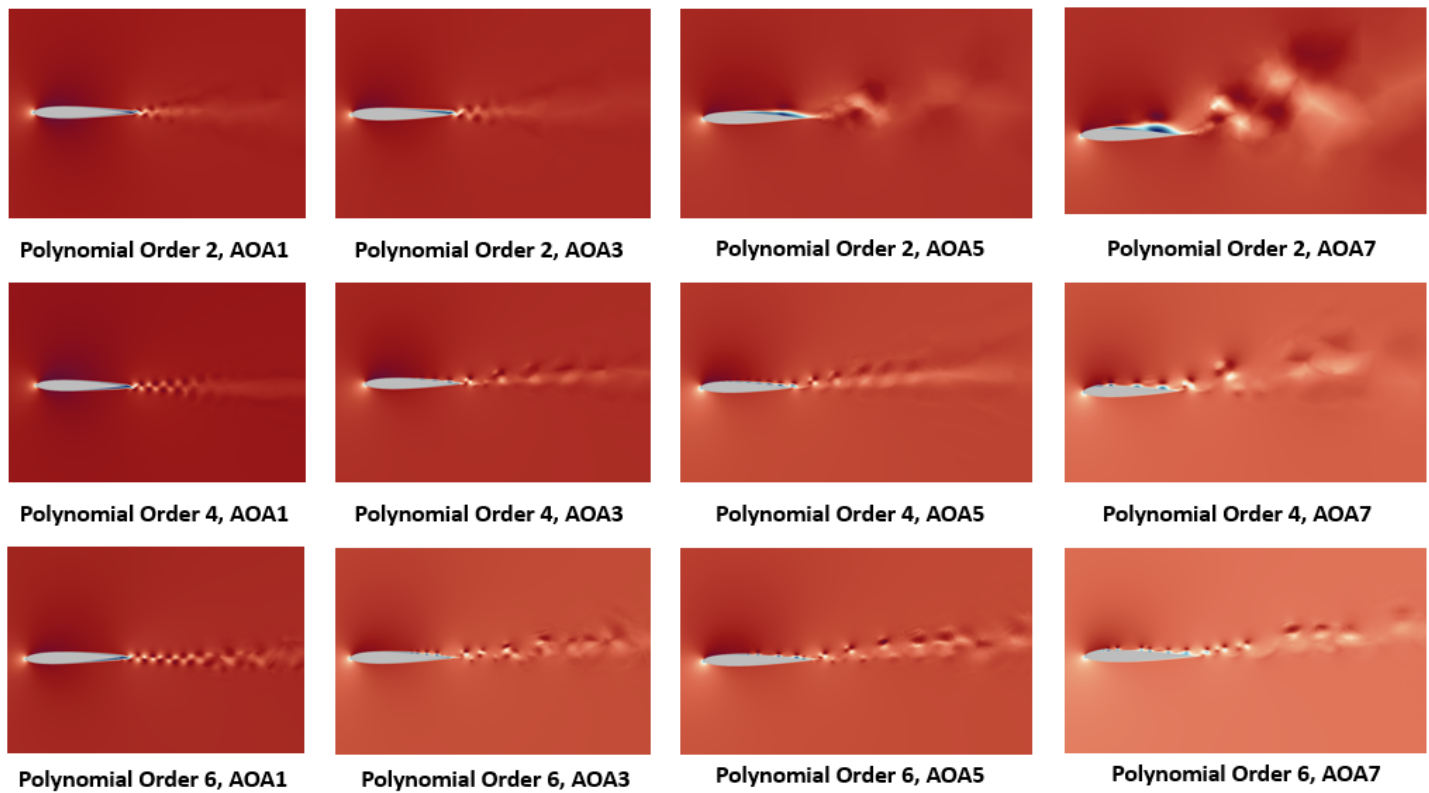

Figure 3.3: Simulation results for polynomial order 6 with respect to different angle of attack 1°, 3°, 5° and 7° for 1 million iteration steps

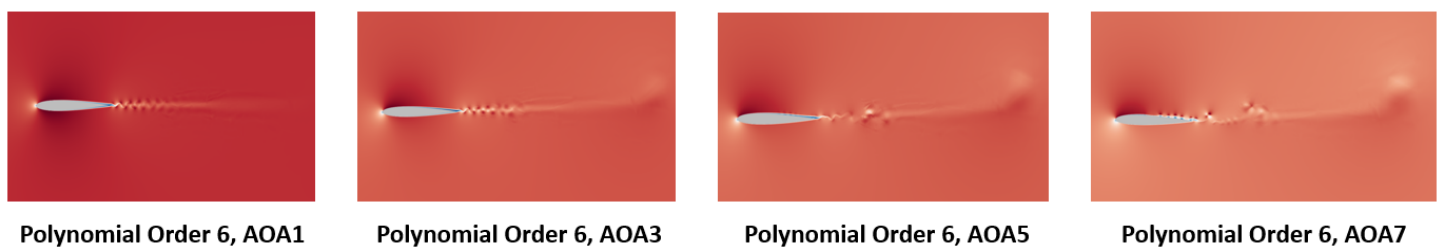

Figure 3.4: Simulation results for polynomial order 6 with respect to different angle of attack 1°, 3°, 5° and 7° for 0.3 million iteration steps

## 3.2 Low-Fidelity Simulation

Low-fidelity simulation has been performed using XFOIL package. Fig 3.5 represents the XFOIL simulation data for the NACA0012 airfoil.

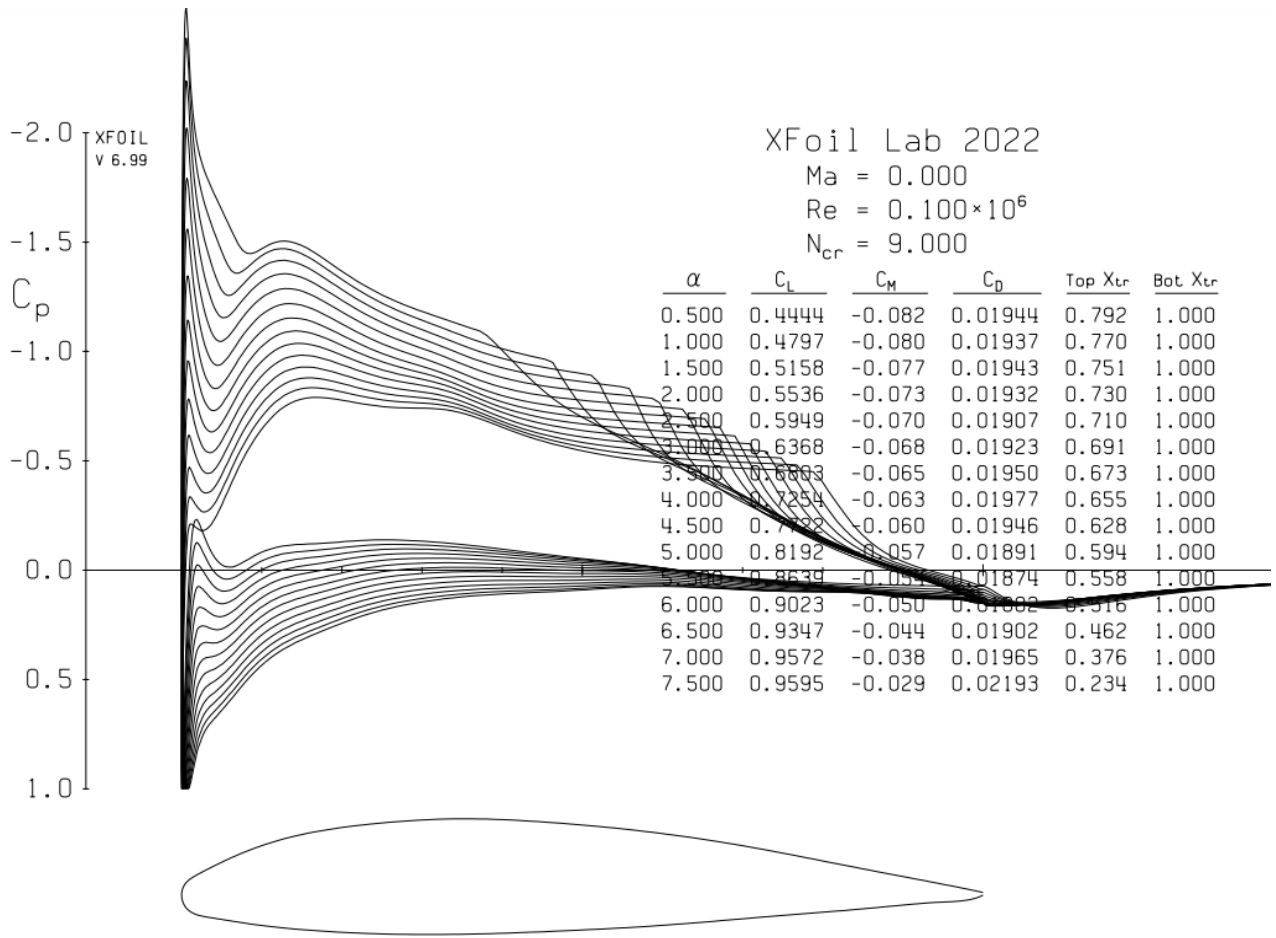

Figure 3.5: XFOIL simulation for NACA0012

## 3.3 Co-Kriging Data Fusion

Co-Kriging data fusion results shown in the Figure 3.6 demonstrates the benefit of integrating low and high fidelity CFD simulations by developing the precise meta-model. This strategy is crucial in the industrial setting as high-fidelity models are used for the final validation purpose. The Co-kriging data fusion results have been obtained using Python.

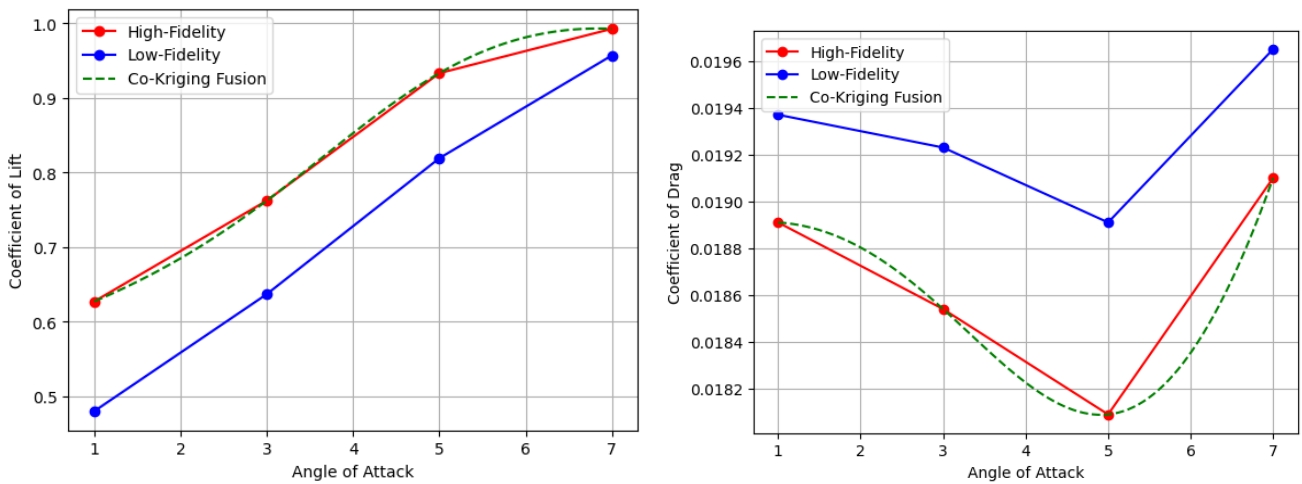

Figure 3.6: Co-kriging data fusion for (a) Lift (left) and (b) Drag (right)

## 3.4 Co-Kriging Adaptive Sampling

The Co-kriging adaptive sampling plot has predicted the unknown values within the specified range as shown in Figure 3.7. This approach is valuable in gaining insights on the coefficient of lift and drag data across that range, eliminating the necessity for additional expensive simulations. The Co-kriging adaptive sampling results have been obtained using Python.

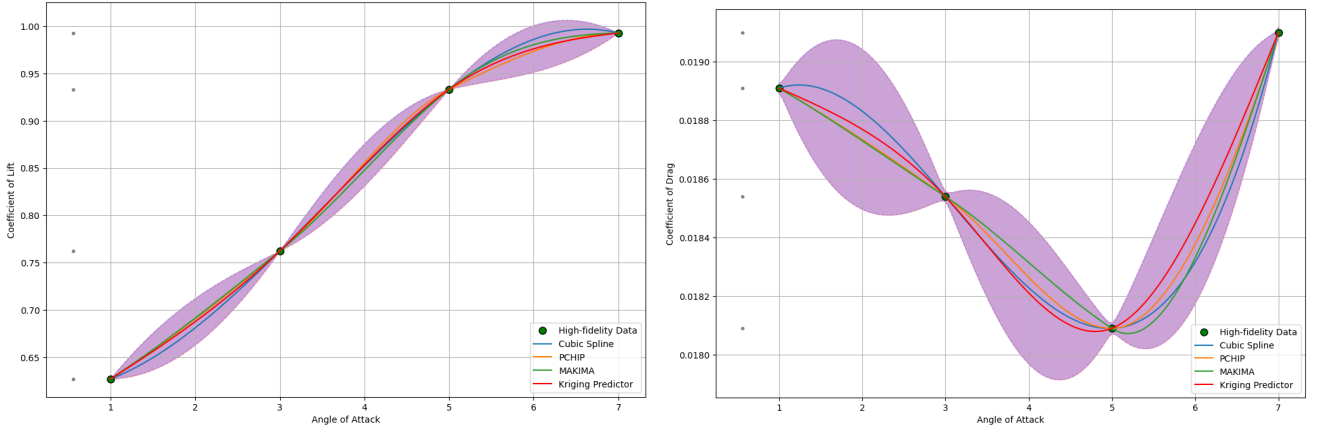

Figure 3.7: Co-kriging adaptive sampling for (a) Lift (left) and (b) Drag (right)

## 3.5 Test Case for Co-Kriging

The **_function test 1_** has been performed with 21 Low-fidelity points and 4 High-fidelity points for the following function:

$$y_L(x) = 0.5y_H + 10(x - 0.5) - 5 \tag{3.1}$$

$$y_H(x) = (6x - 2)^2 \sin(12x - 4) \tag{3.2}$$

Inexpensive data samples are obtained within the interval [0,1] and the high-cost data at $\xi_e = \{0, 0.35, 0.65, 1\}$. The same is shown in the result, Figure 3.8.

The **_function test 2_** has been performed with 21 Low-fidelity points and 6 High-fidelity points for the following function:

$$y_L(x) = 0.5(6x - 2)^2 \sin(12x - 4) + 10(x - 0.5) - 5 \tag{3.3}$$

$$y_H(x) = 0.1y_L(x)^2 + 10 \tag{3.4}$$

Inexpensive data samples are obtained within the interval [0,1] and the high-cost data at $\xi_e = \{0, 0.2, 0.4, 0.6, 0.8, 1\}$. The same is shown in the result, Figure 3.8.

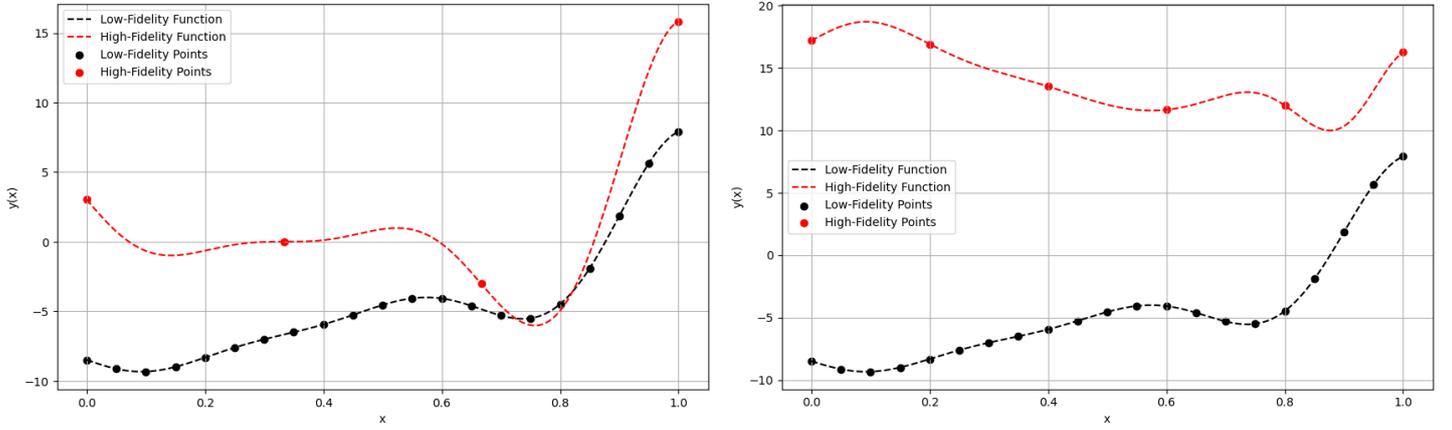

Figure 3.8: (a) Function benchmark test result for equation 3.1 and 3.2 with 21 low-fidelity points and 4 high-fidelity points (left) (b) Function benchmark test result for equation 3.3 and 3.4 with 21 low-fidelity points and 6 high-fidelity points (right)

## 3.6 Surrogate Modelling and Uncertainty Quantification

Several benchmark tests were conducted to observe the performance of the developed multi-fidelity deep neural network model in surrogate modelling and uncertainty quantification.

### 3.6.1 1-dimension function with linear correlation

One dimension function featured by the linear correlation between the LF- and HF data has been tested. The theoretical expression of the function are as follows [16]:

$$y_L(x) = 0.5y_H + 10(x - 0.5) - 5 \tag{3.5}$$

$$y_H(x) = (6x - 2)^2 \sin(12x - 4) \tag{3.6}$$

For the approximation of low-fidelity and high-fidelity functions, 21 low-fidelity points were generated uniformly within the interval [0,1] and 4 high-fidelity points at the coordinates [0.35, 0.75, 1] were generated and then collected as the training data for multi-fidelity deep neural network [16]. Before commencing the training of multi-fidelity deep neural network, the learning rate, number of hidden layers, and number of neurons in each hidden layer were tuned utilizing the Bayesian optimization [16].

The variation range for the hidden layer number spanned from 1 to 4, the variation range for the neuron number in each hidden layer is {8, 16, 24, 32, 40, 48, 56, 64}, and the learning rates were set as {0.01, 0.001, 0.0001} respectively [16]. The optimization objective is to minimize the prediction loss of multi-fidelity deep neural network [16]. The optimal architecture for the low-fidelity deep neural network is structured with three hidden layers comprising 64, 64, and 40 neurons respectively, and the correction deep neural network is finalized with a singular hidden layer with 8 neurons, and the learning rate is 0.001 [16].

From figure 3.9, it is evident, multi-fidelity deep neural network can approximate the high-fidelity function precisely based on 4 high-fidelity data points. This indicates that the developed Correction deep neural network with the ReLU activation function can approximate the linear correlation between the low-fidelity and high-fidelity data [16].

In addition to that, prediction results obtained from the proposed multi-fidelity deep neural network model were compared with Radial basis function and Co-Kriging model, Refer Table 3.2 [16]. The performance of the MF-DNN model is better in comparison to other surrogate models, except in comparison to the Co-kriging model [16]. The Co-kriging model was originally developed keeping the low-dimensional linear assumptions [16]. This shows the effectiveness of the neural networks with the ReLU activation function in approximating the linear correlation between the low-fidelity and high-fidelity data [16].

Table 3.2: Mean Square Error (MSE) prediction for function benchmark test

| Model | MSE Value |
|---|---|
| MF-DNN ReLU | 9.73e-03 |
| RBF | 28.5204e+00 |
| Co-Kriging | 4.77336e-09 |

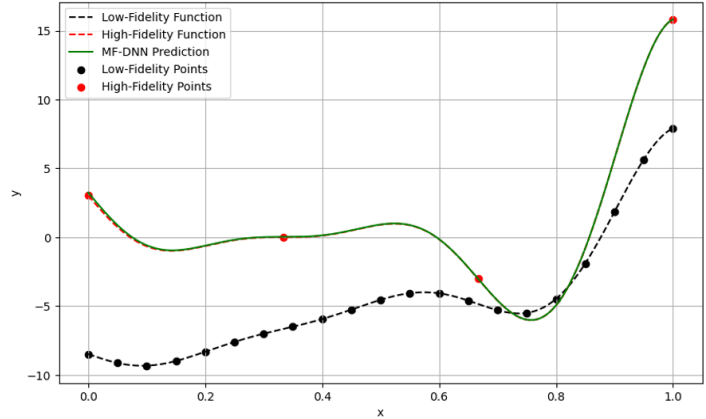

Figure 3.9: MF-DNN in approximating 1-dimension function

### 3.6.2 1-dimension function with non-linear correlation

The one-dimension function featured by the non-linear correlation between the low-fidelity and high-fidelity data has been tested. The theoretical expressions of the function are as follows:

$$y_L(x) = 0.5(6x - 2)^2 \sin(12x - 4) + 10(x - 0.5) - 5 \tag{3.7}$$

$$y_H(x) = 0.1 y_L(x)^2 + 10 \tag{3.8}$$

For the approximation of Low-Fidelity and High-Fidelity functions, 21 low-fidelity points were generated uniformly within the interval [0,1] and 6 high-fidelity points at locations [0, 0.2, 0.4, 0.6, 0.8, 1] were generated and then collected as the training data for multi-fidelity deep neural network [16]. The "ReLU" activation function has been used to test the capability in approximating the non-linear correlation between the low-fidelity and high-fidelity function [16]. The hyperparameters of the neural networks were fine-tuned using the Bayesian optimization, with a training of 1800 epochs, resulting in the low-fidelity deep neural network being configured with three hidden layers, the first two with 64 neurons each, and the third with 40 neurons [16]. In parallel, the correction deep neural network was optimized to two hidden layers with 64 and 56 neurons, respectively [16].

As shown in Figure 3.10, the multi-fidelity deep neural network is accurately approximating the high-fidelity function based on 6 data points [16]. In addition to that, multi-fidelity deep neural network performance was compared against other models, Radial basis function and Co-Kriging. The multi-fidelity deep neural network model has better prediction accuracy as compared to the other models. This demonstrates the effectiveness of the neural network model with the ReLU activation function. It can effectively approximate the non-linear correlation between the low-fidelity and high-fidelity data [16].

Table 3.3: Mean Square Error (MSE) prediction for the function benchmark test

| Model | MSE Value |
|---|---|
| MF-DNN ReLU | 1.17e-02 |
| RBF | 1.4876e+00 |
| Co-Kriging | 4.4034435e+00 |

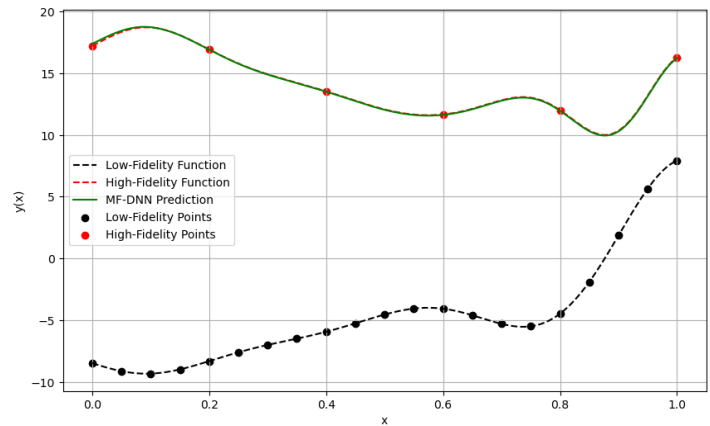

Figure 3.10: MF-DNN in approximating 1-dimension function

### 3.6.3 32-dimensional function

The capability of the Multi-Fidelity Deep Neural Network model in approximating the 32-dimensional function has been tested. The theoretical expressions of the function is shown below:

$$y_L(x_0, \ldots, x_{31}) = 0.8 \times y_H - 0.4 \sum_{i=0}^{30}(x_i x_{i+1}) - 50, \quad x_i \in [-3, 3] \tag{3.9}$$

$$y_H(x_0, \ldots, x_{31}) = (x_0 - 1)^2 + \sum_{i=1}^{31}(2x_i^2 - x_{i-1}^2)^2, \quad x_i \in [-3, 3] \tag{3.10}$$

In order to approximate the high-dimensional function, a substantial dataset consisting of 200,000 low-fidelity and 2,000 high-fidelity data points has been generated using the Latin hypercube sampling method [16]. The low-fidelity deep neural network is configured with two hidden layers, one comprising of 512 neurons and other comprising of 256 neurons, and the correction deep neural network with one layer of 32 neurons. The ReLU activation function has been used for the correction deep neural network with a learning rate of 0.001 [16]

The weights and biases of the multi-fidelity deep neural network were adjusted using the ADAM optimizer for the first 400 iterations [16]. Figure 3.11 shows the multi-fidelity deep neural network prediction in approximating 32-dimension function. Here, the $x$ and $y$ axes represent the multi-fidelity deep neural network predictions and the analytical solutions, respectively. The alignment of the red points with the analytical solutions underscores the model's accuracy in solving high-dimensional problem [16]. The accuracy of multi-fidelity deep neural network model is good in solving 32-dimension function, and Co-kriging model is unable to yield results for this problem because of the high memory requirement [16].

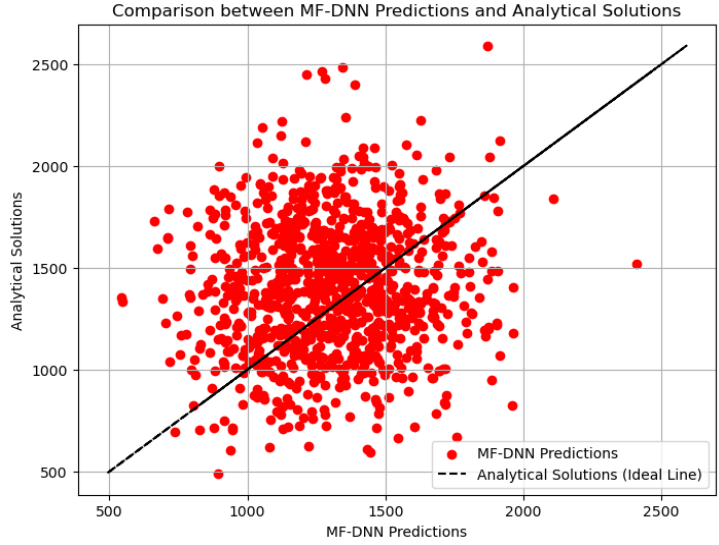

Figure 3.11: MF-DNN in approximating 32-dimensional function

### 3.6.4 100-dimensional function

The performance of the proposed multi-fidelity deep neural network model has been evaluated by testing 100-dimensional benchmark function. The equation has been expressed in the following form as shown below:

$$y_L(x_0, \ldots, x_{99}) = 0.8 \times y_H - 0.4 \sum_{i=0}^{98}(x_i x_{i+1}) - 50, \quad x_i \in [-3, 3] \tag{3.11}$$

$$y_H(x_0, \ldots, x_{99}) = (x_0 - 1)^2 + \sum_{i=1}^{99}(2x_i^2 - x_{i-1}^2)^2, \quad x_i \in [-3, 3] \tag{3.12}$$

The proposed multi-fidelity deep neural network model was trained using the 10,000,000 low fidelity data points and 100,000 high fidelity data points [16]. The architecture of the low-fidelity deep neural network included four layers with 512, 512, 256, and 128 neurons respectively, while the high-fidelity deep neural network was constructed with a singular layer of 64 neurons [16]. The ReLU activation function was implemented in both the networks with a learning rate of 0.001 [16]. The entire training process was carried out in the TensorFlow environment on *Google Colaboratory* [16].

Multi-fidelity deep neural network prediction error should be minimal and it is indicated by the red scatter points which aligns closely with the black line, which denotes analytical solution line [16]. Thus, the alignment of the red points with the analytical solutions underscores the model's accuracy in solving high-dimensional problem [16]. From figure 3.12, it is evident, multi-fidelity deep neural network has demonstrated its ability to predict the analytical solution of 100-Dimension function effectively. It is also evident, the multi-fidelity deep neural network model is effective in addressing high-dimensional surrogate modelling problems [16]. It is also noted that the Radial Basis function, Kriging and Co-Kriging models has experienced difficulties during the training process due to extensive memory requirements leading to an inability to complete the training process successfully [16].

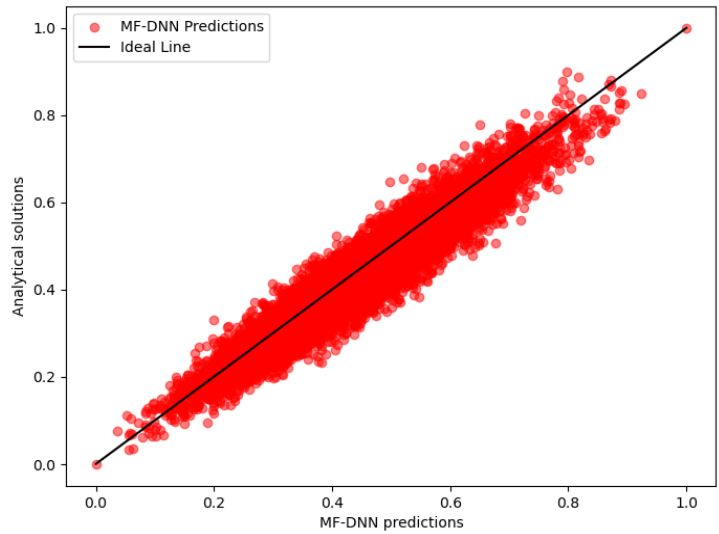

Figure 3.12: Performance of MF-DNN model for 100-dimensional function

### 3.6.5 Uncertainty Quantification for 1-Dimension function

The assessment of input aleatory uncertainties on the Quality of Interest has been conducted using the Multi-Fidelity Deep Neural Network model in combination with the Monte Carlo sampling technique [16]. The governing equations, 3.5 and 3.6 were applied for uncertainty propagation, focusing on two main input uncertainty distributions: uniform and Gaussian. The input $x$ within the domain $\Gamma$ was modeled as a uniform random variable on $\Gamma = U[0.6, 0.8]$ or as a Gaussian random variable on $\Gamma = N[0.7, 0.03^2]$ [16]. To generate the Gaussian distributions within the specific bounded interval, multidimensional truncated Gaussian method was utilized because of its effectiveness [16].

The histogram comparison of the Quality of Interest probability density under varying input uncertainty distributions is shown in Figure 3.13 [16]. The results of this comparison shows that the peak of the QoI probability is around -5.5 for both uniform and Gaussian input uncertainties [16]. A shift is also observed in the QoI's probability density, which tends to cluster in the range of [-5.5, -5.0], when the input uncertainty shifts from a uniform to a Gaussian distribution [16]. This overall tendency demonstrates that Multi-fidelity deep neural network can accurately simulate the propagation of uncertainty in systems with lower dimensions [16].

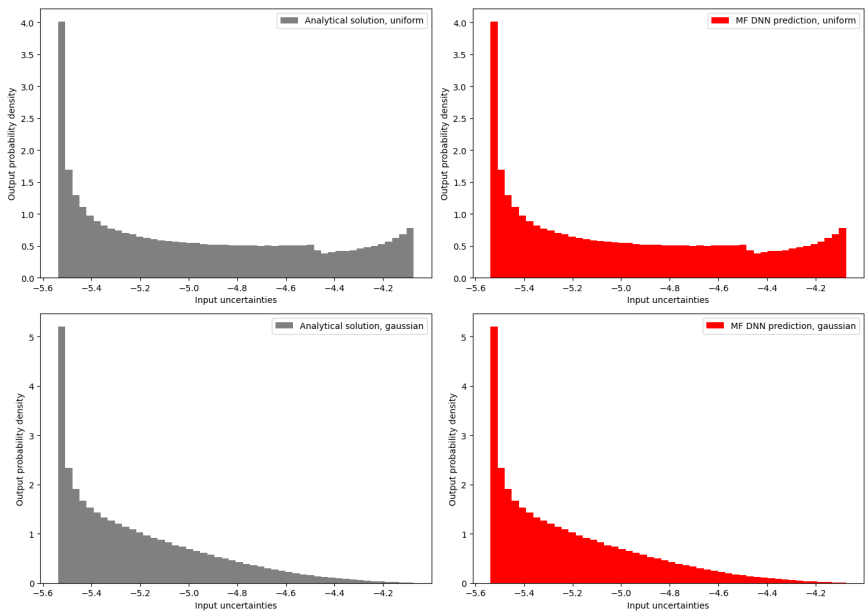

Figure 3.13: Histogram comparison of QoI probability density distribution, 1-dimension function

### 3.6.6 Uncertainty Quantification for 32-Dimensional function

In order to evaluate the proficiency of the Multi-Fidelity Deep Neural Network model in high-dimensional uncertainty quantification scenarios, a benchmark function encompassing 32 dimensions, was utilized to dictate the propagation of uncertainty [16]. The input variable $x$, confined to the domain $\Gamma$, was modeled as either a 32-dimensional uniform random variable over the domain $\Gamma = U[-3, 3]$, or as a Gaussian random variable over $\Gamma = N(0, 1^2)$. The generation of the Gaussian distributions across the 32 dimensions was achieved using the multidimensional truncated Gaussian technique, and Monte Carlo simulation was adopted to emulate the uncertainty propagation[16].

The histogram comparison of the Quality of Interest probability density under varying input uncertainty distributions for the 32-dimensional function is shown in Figure 3.14 [16]. It is observed that the QoI's probability density functions exhibit leftward skew as the input uncertainties transition from a uniform to a Gaussian distribution [16]. The overall tendancy demonstrates that multi-fidelity deep neural network can effectively capture both the qualitative and quantitative aspects of the uncertainty propagation in complex, high-dimensional UQ problems[16].

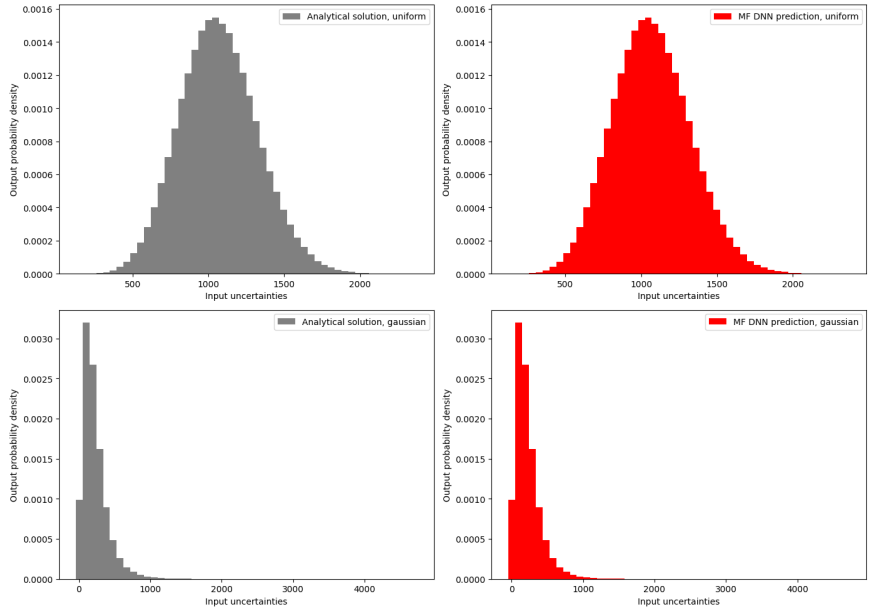

Figure 3.14: Histogram comparison of QoI probability density distribution, 32-dimensional function

### 3.6.7 Uncertainty Quantification for 100-Dimensional function

The Multi-Fidelity deep neural network has been deployed to address the challenge posed by a 100-dimensional uncertainty quantification (UQ) problem. Input uncertainty, represented as $x \in \Gamma$, took on two variants: a 100-dimensional uniform distribution across $\Gamma = U[-1, 1]$, or a 100-dimensional Gaussian distribution $\Gamma = N(0, 1^2)$ [16]. In addition to that, multi-fidelity deep neural network is used to deduce the probability density distributions of the QoI, the comparison is shown in Figure 3.15. The probability distributions obtained from the multi-fidelity deep neural network closely mirrors with the analytical solutions, indicating an effective demonstration of the overall data distribution shape and characteristics [16].

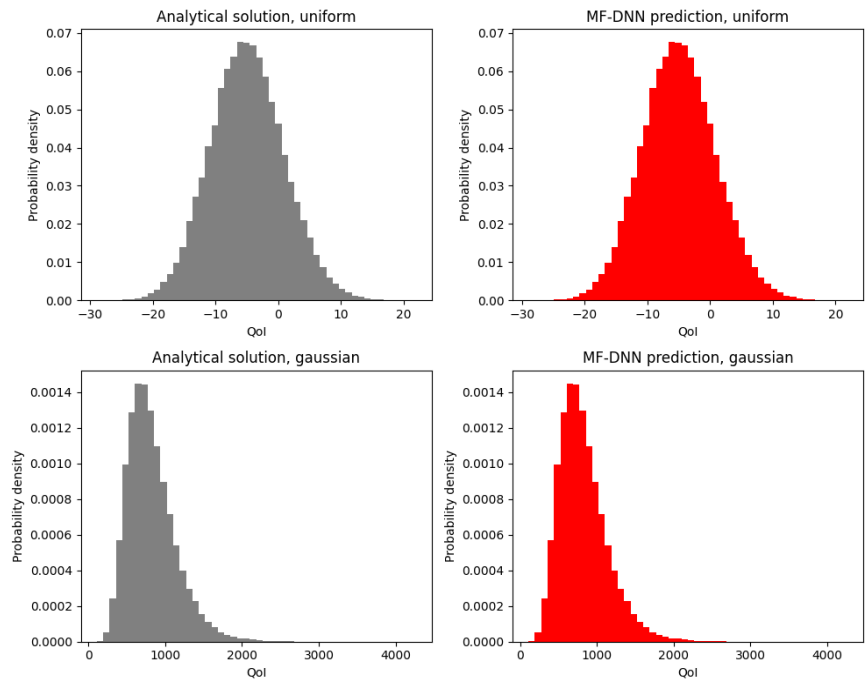

Figure 3.15: Histogram comparison of QoI probability density distribution, 100-dimensional function

# Chapter 4

# Conclusion and Future work

The Uncertainty Quantification research primarily focuses on gathering the high-fidelity and low-fidelity simulation data using Nektar++ and XFOIL package respectively. Further, Co-kriging techhnique has been used to obtain the precise data predictions for the lift and drag within the confined domain without conducting the costly simulations on HPC clusters. To further minimize the reliability on high-fidelity numerical simulations in Uncertainty Quantification, a multi-fidelity strategy has been adopted. The core principle behind the multi-fidelity method is to maintain the accuracy of the surrogate model predictions by leveraging a large volume of low-fidelity data complemented by a smaller set of high-fidelity data [16]. The Multi-fidelity Deep Neural Network is particularly chosen because of its capability in addressing surrogate modeling and UQ challenges. Also, the deep neural network inherent ability to approximate functions universally. In addition to that, error analysis has been conducted to assess the performance of both Co-kriging and Multi-fidelity deep neural network model, revealing that the Multi-fidelity deep neural network model has outperformed Co-kriging in effectiveness [16].

The effectiveness of the multi-fidelity deep neural network has been validated through the approximation of benchmark functions across 1-, 32-, and 100-dimensional spaces, encompassing both linear and non-linear correlations [16]. The surrogating modelling results showed that multi-fidelity deep neural network has shown excellent approximation capabilities for the test functions [16]. Further applications involved utilizing the multi-fidelity deep neural network for the simulation of aleatory uncertainty propagation in 1-, 32-, and 100-dimensional function test, considering both uniform and Gaussian distributions for input uncertainties [16]. The results have shown that multi-fidelity deep neural network model has efficiently predicted the probability density distributions of quantities of interest (QoI) as well as the statistical moments with precision and accuracy [16]. The Co-Kriging model has exhibited limitations when addressing 32-Dimension problems due to the lack of memory capacity for storage and manipulation [16]. In addition to that, Radial Basis Function, Kriging, and Co-Kriging models also posses computational challenges for 100-Dimension function, because of the required system memory for training[16].

From the computational fluid dynamic research, it has also been observed, higher polynomial order yields a more accurate representation of the airfoil surface because it encompasses more terms, allowing for a better fit to the data points. Consequently, the coefficient of lift calculated from the higher polynomial distribution is more precise. This is the reason why the lift and drag data obtained from the polynomial order 6 has been used for further research investigation as compared to the data obtained from polynomial order 2 and 4. Initially, computational fluid dynamics data were obtained for the polynomial order of 6, for different angles of attack over 0.3 million steps. However, the initial results did not exhibit convergence and revealed discrepancies in the flow visualization. Due to this reason, the simulations were performed for 1 million steps to address these issues.

The research creates a methodology for quantifying uncertainty in computational fluid dynamics with a focus to minimize the required number of samples. This significantly lowers the computational costs [15]. The research also shows the the benefits of integrating low and high fidelity CFD simulations by developing precise meta-model[15].

The choice for the CFD investigation centered on the NACA0012 airfoil, primarily due to its ready availability of experimental data and its extensive utilization across the aerospace industry. Future research endeavors encompass 3-dimensional Large Eddy & Direct Numerical Simulations at higher angles of attack for NACA airfoil (Fig. 4.1).

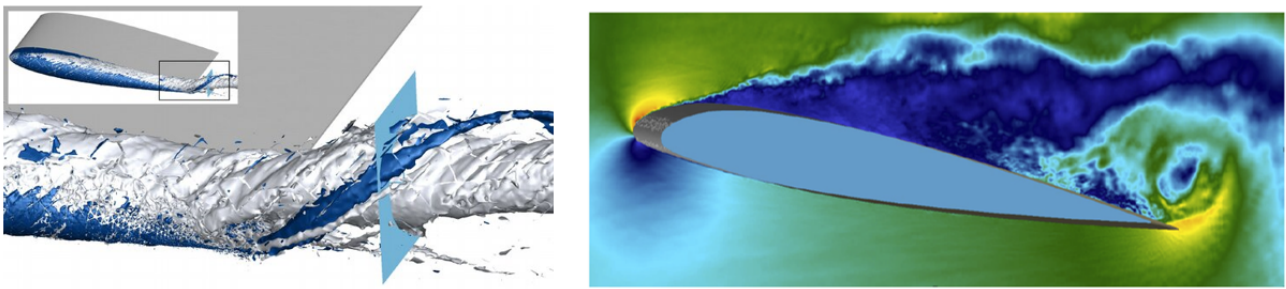

Figure 4.1: High-fidelity 3-Dimensional simulation from [2], [18]

Another avenue of future study involves employing data-centric strategies in space exploration to meet the objective of achieving a net-zero emissions target through the use of high-fidelity simulation data. The intensive demands of detailed computational fluid dynamics simulations make data-driven methods more appealing, as it produces more precise and reliable predictions. Electric aircraft exhibit operational characteristics that differ from those of traditional aircraft. Custom-designed airfoils can enhance the efficiency of electric propulsion systems, thereby further lowering emissions. This strategy can significantly improve aerodynamic efficiency, leading to reductions in the usage of the fuel and emissions. Meeting the ambitious goal of net-zero emissions by 2030 presents a substantial global challenge. This research can significantly contribute to the sustainability objectives of the industry by offering viable solutions to enhance fuel efficiency. Consequently, this will help in decreasing the aviation sector's carbon footprint, aligning with worldwide efforts to address the climate change.

# Bibliography

[1] A Bolis, CD Cantwell, RM Kirby, and SJ Sherwin. From h to p efficiently: optimal implementation strategies for explicit time-dependent problems using the spectral/hp element method. *International journal for numerical methods in fluids*, 75(8):591–607, 2014.

[2] B Bornhoft, SS Jain, K Goc, ST Bose, and P Moin. Wall-modeled les of laser-scanned rime, glaze, and horn ice shapes. *Annual Research Briefs, Center for Turbulence Research*, pages 71–85, 2022.

[3] Brett Bornhoft, Suhas Jain, Konrad Goc, Sanjeeb Bose, and Parviz Moin. Large-eddy simulation of a naca23012 airfoil under clean and iced conditions. Technical report, SAE Technical Paper, 2023.

[4] Chris D Cantwell, David Moxey, Andrew Comerford, Alessandro Bolis, Gabriele Rocco, Gianmarco Mengaldo, Daniele De Grazia, Sergey Yakovlev, J-E Lombard, Dirk Ekelschot, et al. Nektar++: An open-source spectral/hp element framework. *Computer physics communications*, 192:205–219, 2015.

[5] Richard Courant. Variational methods for the solution of problems of equilibrium and vibrations. 1943.

[6] Gyorgy Farkas. Development of a python interface to nektar++. Technical report, Imperial College London, 2020.

[7] Bruce A Finlayson. *The method of weighted residuals and variational principles*. SIAM, 2013.

[8] Alexander Forrester, Andras Sobester, and Andy Keane. *Engineering design via surrogate modelling: a practical guide*. John Wiley & Sons, 2008.

[9] Alexander IJ Forrester, András Sóbester, and Andy J Keane. Multi-fidelity optimization via surrogate modelling. *Proceedings of the royal society a: mathematical, physical and engineering sciences*, 463(2088):3251–3269, 2007.

[10] Boris Grigoryevich Galerkin. Series solution of some problems of elastic equilibrium of rods and plates. *Vestnik inzhenerov i tekhnikov*, 19(7):897–908, 1915.

[11] Walid Hambli, James Slaughter, Filipe Fabian Buscariolo, and Spencer Sherwin. Extension of spectral/hp element methods towards robust large-eddy simulation of industrial automotive geometries. *Fluids*, 7(3):106, 2022.

[12] Alexander Hrennikoff. Solution of problems of elasticity by the framework method. 1941.

[13] Emilia Juda. Development of a python interface to nektar++. Technical report, Imperial College London, 2018.

[14] George Karniadakis and Spencer J Sherwin. *Spectral/hp element methods for computational fluid dynamics*. Oxford University Press, USA, 2005.

[15] Kerry Klemmer. Kriging and co-kriging with adaptive sampling for uncertainty quantification in computational fluid dynamics. Technical report, Imperial College London, 2015.

[16] Zhihui Li and Francesco Montomoli. Surrogate modelling and uncertainty quantification based on multi-fidelity deep neural network. *arXiv preprint arXiv:2308.01261*, 2023.

[17] Zhihui Li, Francesco Montomoli, Nicola Casari, and Michele Pinelli. High-dimensional uncertainty quantification of high-pressure turbine vane based on multifidelity deep neural networks. *Journal of Turbomachinery*, 145(11), 2023.

[18] Jean-Eloi W Lombard, David Moxey, Spencer J Sherwin, Julien FA Hoessler, Sridar Dhandapani, and Mark J Taylor. Implicit large-eddy simulation of a wingtip vortex. *AIAA Journal*, 54(2):506–518, 2016.

[19] Xuhui Meng and George Em Karniadakis. A composite neural network that learns from multi-fidelity data: Application to function approximation and inverse pde problems. *Journal of Computational Physics*, 401:109020, 2020.

[20] Francesco Montomoli, Mauro Carnevale, Antonio D'Ammaro, Michela Massini, and Simone Salvadori. *Uncertainty quantification in computational fluid dynamics and aircraft engines*. Springer, 2015.

[21] Mohammad Motamed. A multi-fidelity neural network surrogate sampling method for uncertainty quantification. *International Journal for Uncertainty Quantification*, 10(4), 2020.

[22] Alejandro Iglesias Perez. Quasi 3d computation of the taylor-green vortex flow using spectral/hp element methods. Technical report, Imperial College London, 2016.

[23] Apostolos F Psaros, Xuhui Meng, Zongren Zou, Ling Guo, and George Em Karniadakis. Uncertainty quantification in scientific machine learning: Methods, metrics, and comparisons. *Journal of Computational Physics*, 477:111902, 2023.

[24] Chris D. Cantwell Robert M. (Mike) Kirby, Spencer J. Sherwin and David Moxey. Nektar++: Developer guide, https://www.nektar.info/src/developer-guide-5.3.0.pdf. Technical report, Imperial College London, Accesssed September 2023.

[25] Eiji Sakai, Meng Bai, Richard Ahlfeld, Kerry Klemmer, and Francesco Montomoli. Bi-fidelity uq with combination of co-kriging and arbitrary polynomial chaos: Film cooling with back facing step using rans and des. *International Journal of Heat and Mass Transfer*, 131:261–272, 2019.

[26] Spencer J Sherwin. Spencer sherwin. nektar++ website gallery. technical report. Accessed March 2024.

[27] Alexandre Sidot. Spectral/hp element methods applied to vortex structures in a low incidence delta wing wake compared to experiments. Technical report, Imperial College London, 2017.

[28] Maksym Tymchenko. Development of a python interface to nektar++. Technical report, Imperial College London, 2019.

[29] Qiqi Wang, Tonkid Chantrasmi, Gianluca Iaccarino, and Parviz Moin. Adaptive uncertainty quantification using adjoint method and generalized polynomial chaos. In *APS Division of Fluid Dynamics Meeting Abstracts*, volume 59, pages EM–008, 2006.

[30] Hui Xu, Chris D Cantwell, Carlos Monteserin, Claes Eskilsson, Allan P Engsig-Karup, and Spencer J Sherwin. Spectral/hp element methods: Recent developments, applications, and perspectives. *Journal of Hydrodynamics*, 30:1–22, 2018.

[31] Zhen-Guo Yan, Yu Pan, Giacomo Castiglioni, Koen Hillewaert, Joaquim Peiró, David Moxey, and Spencer J Sherwin. Nektar++: Design and implementation of an implicit, spectral/hp element, compressible flow solver using a jacobian-free newton krylov approach. *Computers & Mathematics with Applications*, 81:351–372, 2021.

[32] Xinshuai Zhang, Fangfang Xie, Tingwei Ji, Zaoxu Zhu, and Yao Zheng. Multi-fidelity deep neural network surrogate model for aerodynamic shape optimization. *Computer Methods in Applied Mechanics and Engineering*, 373:113485, 2021.

# Appendix A

# Structure of Report

The structure of the UQ report is as follows: It begins with an Abstract, followed by Chapter 1, which discusses about the background and literature review related to the research. Chapter 2 details the methodology, covering the introduction to Nektar++, the Nektar++ libraries, and the simulation process for acquiring high-fidelity data, as well as the XFOIL simulation for acquiring low-fidelity data. It also details about Co-kriging data fusion, Co-kriging Adaptive Sampling, and UQ through Surrogate Modeling, with a particular focus on the Multi-fidelity Deep Neural Network approach. Chapter 3 presents the results and discussions, discussing about the high-fidelity and low-fidelity simulations results, results from Co-kriging data fusion and Adaptive Sampling, and Test case results using Co-kriging and Test case comparison results using Multi-fidelity Deep Neural Network (MF-DNN). The final chapter concludes the report and outlines future directions, providing a thorough analysis of the conducted research and potential for extending the current work.

# Appendix B

# Computational Fluid Dynamics

Commercial CFD software packages use a variety of numerical solvers, including finite difference, finite volume, and finite element solvers. Finite difference methods (FDMs) represent the most established numerical techniques for Computational Fluid Dynamics, involving the discretization of the governing equations in both space and time through a grid of points. Implementing FDMs is straightforward and highly efficient when dealing with uncomplicated geometries, but their application becomes challenging when addressing complex geometries, potentially leading to reduced accuracy compared to other methods.

### B.0.1 Finite difference methods

Finite difference method are obtained using the definition of the first derivative of the function u$_x$[13]

The right-hand side of the equation becomes the exact solution, when $\Delta x \to 0$. $u_x(x) = \frac{\partial u}{\partial x}(x) = \lim_{\Delta x \to 0} \frac{u(x+\Delta x)-u(x)}{\Delta x}$ (B.1)

Finite difference methods evaluate the function value $u_i = u(x_i)$ at discrete points $x_i$ on the constructed grid. Forward difference scheme can be expressed in the following form:

$$\frac{\partial u}{\partial x}(x_i) = \frac{u_{i+1} - u_i}{\Delta x} \tag{B.2}$$

Backward difference scheme (equation B.3) and Central difference scheme (equation B.4) can be expressed in the following form respectively:

$$\frac{\partial u}{\partial x}(x_i) = \frac{u_i - u_{i-1}}{\Delta x} \tag{B.3}$$

$$\frac{\partial u}{\partial x}(x_i) = \frac{u_{i+1} - u_{i-1}}{2\Delta x} \tag{B.4}$$

## B.0.2 Finite Volume method

Finite volume methods (FVMs), although akin to FDMs, adopt a distinct approach to discretize governing equations. They rely on the principle of conservation, making them highly suitable for addressing complex geometry problems. FVMs are known for their relatively straightforward implementation and efficiency. Nonetheless, it's worth noting that for certain problems, they may not match the accuracy achieved by spectral methods [13]

Finite volume method deals with the differential equation to consider the integral over the region. Consider a general differential equation represented by the linear differential operator L(u) [13]

$$L(u) = q \tag{B.5}$$

The residual of the approximate numerical solution $u^\delta(x)$ is defined as the difference between the approximate solution and the exact solution. Mathematically, the residual is given by:

$$R(u^\delta) = L(u^\delta) - q \tag{B.6}$$

Where $u^\delta(x) = u(x)$, $R(u) = 0$

The method of weighted residuals (MWR) is a technique for solving the partial differential equations (PDEs). It works by multiplying the PDE by a weight function w(x) and integrating it over the domain $\Omega$. This gives rise to a new equation called the weighted residual equation.

$$\int_\Omega w(x) R(u^\delta(x)) dx = \int_\Omega w(x) L(u^\delta(x)) dx - \int_\Omega w(x) q(x) dx \tag{B.7}$$

The weak form of a differential equation is obtained by setting the weighted residual to zero.

$$\int_\Omega w(x) L(u^\delta(x)) dx = \int_\Omega w(x) q(x) dx \tag{B.8}$$

The weight is chosen as following:

$$w_i(x) = \begin{pmatrix} 1 & if x \in \Omega_i \\ 0 & if x \notin \Omega_i \end{pmatrix} \tag{B.9}$$

## B.0.3 Finite Element Method

Finite element method was developed to solve the problem in solid mechanics, and it was later evolved for its implementation in the field of computational fluid dynamics by solving the Euler equation, Navier-Stokes equation, and heat transfer equation.

Hrennikoff first proposed the concept of the finite element. He did this by replacing the continuous material of an elastic body with a framework of bars. This framework could then be used to model a variety of elastic cases, such as the bending of plates and cylindrical shells[12]. Courant[5] was one of the first to discretize the computational domain into triangular regions, which allowed him to solve elliptic partial differential equations. This work laid the foundation for the finite element method, which is now a powerful tool for solving a wide variety of problems in engineering and science.

Following Courant's work, Boris Galerkin[10] developed the Galerkin method, which is a variational approach to solving partial differential equations. The method seeks an approximate solution to the equation that minimizes the residual. The main idea of the method is to use a system of linearly independent test functions to project the residual onto a finite-dimensional space.

Finite element methods (FEMs) are the most flexible numerical methods for CFD. They are used to solve problems with complex geometries and are very accurate.

In the finite element method (FEM), the unknown solution is approximated on the discretized domain by piecewise polynomial functions. The degree of the polynomial functions can be chosen to be low (first-order) or high (higher-order).

Low-order FEM uses linear polynomial functions to approximate the solution. This is the simplest and most common type of FEM. However, it can be inaccurate for problems with sharp features or discontinuities. High-order FEM uses higher-order polynomial functions to approximate the solution. This can achieve higher accuracy than low-order FEM, but it is more computationally expensive. The h-type FEM refines the mesh to improve accuracy. This means that the size of the elements in the mesh is reduced. The degree of the polynomial functions is kept constant. The p-type FEM enhances the polynomial degree to improve accuracy. This means that the degree of the polynomial functions in some or all elements is increased. The mesh size is kept constant. The hp-type FEM combines the ideas of mesh refinement and polynomial degree enhancement. This can achieve the highest accuracy, but it is the most computationally expensive type of FEM.

### B.0.4 Overview of Spectral Methods

An overview of the spectral method approach has been demonstrated in the below mentioned figure (Figure B.1)

### B.0.5 Detailed Explanation to Nektar++ 2D Simulation

High order mesh has been generated to perform the 2D Nektar++ simulation.

Following is the detailed explanation to the session file for 2-Dimensional simulation:

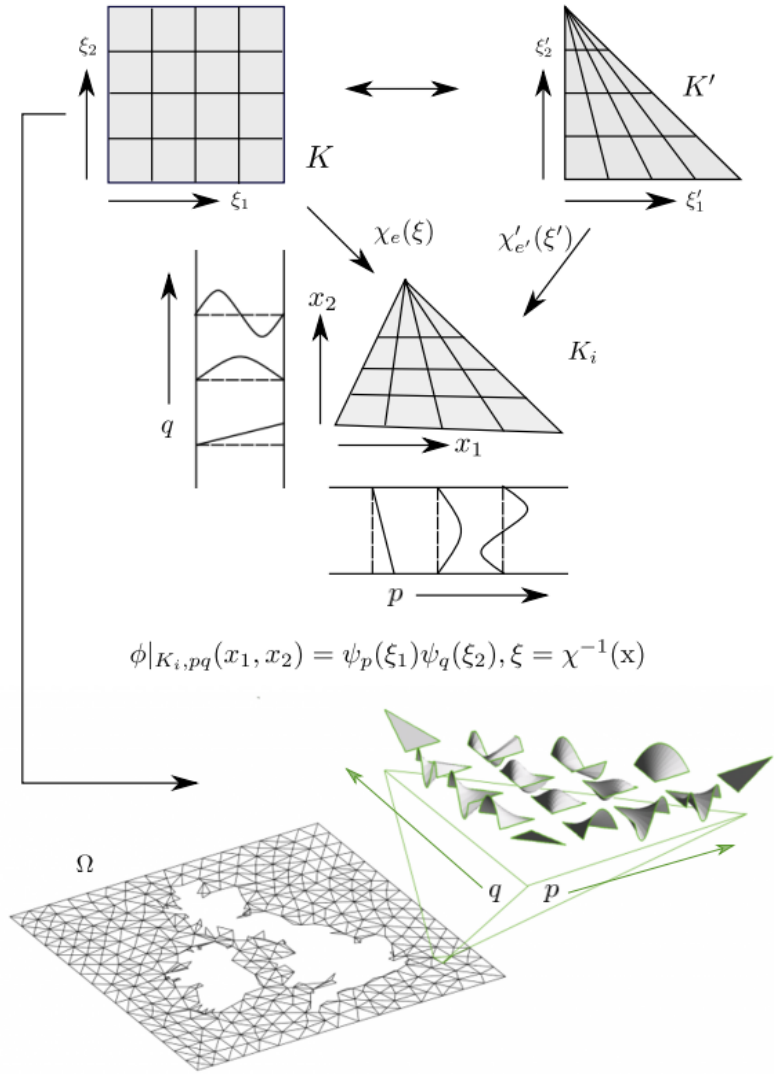

Figure B.1: Construction of 2-Dimension fourth-order C0-continuous modal triangular expansion basis using generalised tensor-product procedure [30]

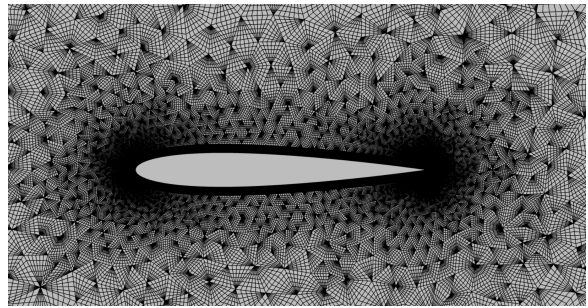

Figure B.2: NACA0012 High-order Mesh

**1. Expansion**

The Expansion line demonstrates the polynomial order of the simulation, where Polynomial Order is equal to NUMMODES – 1, and flow field evaluation has been performed for 3 different NUMMODES (corresponds to 7,5,3).

```
<E COMPOSITE="C[1,2]" NUMMODES="7" TYPE="MODIFIED" FIELDS="u,v" / >

<E COMPOSITE="C[1,2]" NUMMODES="6" TYPE="MODIFIEDQUADPLUS1" FIELDS="p" />
```

## 2. Solver Information

The solver provides an information about the different numerical schemes applied to the Unsteady Incompressible Navier-Stokes equation:

$$\partial u/\partial t = -\nabla p + \nabla^2 u - (u \cdot \nabla)u + f$$

$$\nabla \cdot u = 0$$

```
<I PROPERTY="EQTYPE" VALUE="UnsteadyNavierStokes" / >
```

The solver is solving the Unsteady Navier-Stokes equation.

```
<I PROPERTY="SolverType" VALUE="VelocityCorrectionScheme" / >
```

The velocity and pressure are decoupled under the Velocity correction scheme, so that they are calculated separately. This benefits the computation efficiency but leads to error. The error and the temporal preciseness of the scheme can have the same order of magnitude under the condition; discretised pressure boundary conditions [27]

```
<I PROPERTY="EvolutionOperator" VALUE="Nonlinear"/ >

<I PROPERTY="AdvectionForm" VALUE="Convective"/ >
```

The Convective Advection form is used in the solver model.

```
<I PROPERTY="Projection" VALUE="Galerkin" / >
```

Galerkin Projection is used at the root of spectral/hp element method.

```
<I PROPERTY="TimeIntegrationMethod" VALUE="IMEXOrder2" / >
```

Second order implicit-explicit time-integration scheme is used (This parameter defines the time-integration method applied to the partial differential equations).

```
<I PROPERTY="SpectralVanishingViscosity" VALUE="DGKernel" / >
```

It is a stabilization technique that increases the viscosity of the high-frequency modes in the numerical solution. This is done to smooth out oscillations in the solution, which can lead to improved stability of the simulation.

## 3. Global system Information

This section defines the tolerances required on each flow variable for the iterative solver.

```
<V VAR="u,v,w" >
```

```
<I PROPERTY="GlobalSysSoln" VALUE="IterativeMultiLevelStaticCond" />

<I PROPERTY="Preconditioner" VALUE="Diagonal"/>

<I PROPERTY="IterativeSolverTolerance" VALUE="1e-9"/>

<V VAR="p" >

<I PROPERTY="GlobalSysSoln" VALUE="XxtMultiLevelStaticCond" />
```

To enable parallel execution of the solver, it is necessary to modify the parameter from "DirectStaticCond" since it is only suitable for running the solver in serial. The "GlobalSysSoln" parameter is instrumental in configuring the global linear algebraic system (Ax = b).

## 4. 2D Simulation Parameters

**TimeStep = 1.0e-5**

**NumSteps = 1000000**

**IO_CheckSteps = 1000**, Number of steps after which an output file is written

**IO_InfoSteps = 100** , Number of steps after which information are printed to the screen

**IO_CFLSteps = 100**

**Re = 100000.0**

**Kinvis = 1.0/Re** , Value of kinematic viscosity.

**AoA = (Angle of attack (in degree))*PI/180**

**Uinf = 1.0**

**Uinfx = Uinf*cos(AoA)**

**Uinfy = Uinf*sin(AoA)**

## 5. Boundary Conditions

```
<REGION REF="0"><!-- airfoil wall -->

<D VAR="u" VALUE="0" >

<D VAR="v" VALUE="0" >

<N VAR="p" USERDEFINEDTYPE="H" VALUE="0" >

<REGION REF="1"><!-- Inflow -->
```

```
<D VAR="u" VALUE="Uinfx" />

<D VAR="v" VALUE="Uinfy" />

<N VAR="p" USERDEFINEDTYPE="H" VALUE="0" />
```

**<REGION REF="2"><!-- outflow -->**

```
<N VAR="u" USERDEFINEDTYPE="HOutflow" VALUE="0" />

<N VAR="v" USERDEFINEDTYPE="HOutflow" VALUE="0" />

<D VAR="p" USERDEFINEDTYPE="HOutflow" VALUE="0" />
```

The outlet boundary condition is set as "HOutFlow", which imposes a pressure Dirichlet boundary condition. This is different from the fully developed flow condition.

**<REGION REF="3"><!-- side -->**

```
<D VAR="u" VALUE="Uinfx" />

<D VAR="v" VALUE="0" />

<N VAR="p" USERDEFINEDTYPE="H" VALUE="0" />

</BOUNDARYCONDITIONS>

<FUNCTION NAME="InitialConditions">

<E VAR="u" VALUE="Uinfx"/>

<E VAR="v" VALUE="Uinfy"/>

<E VAR="p" VALUE="0"/>
```

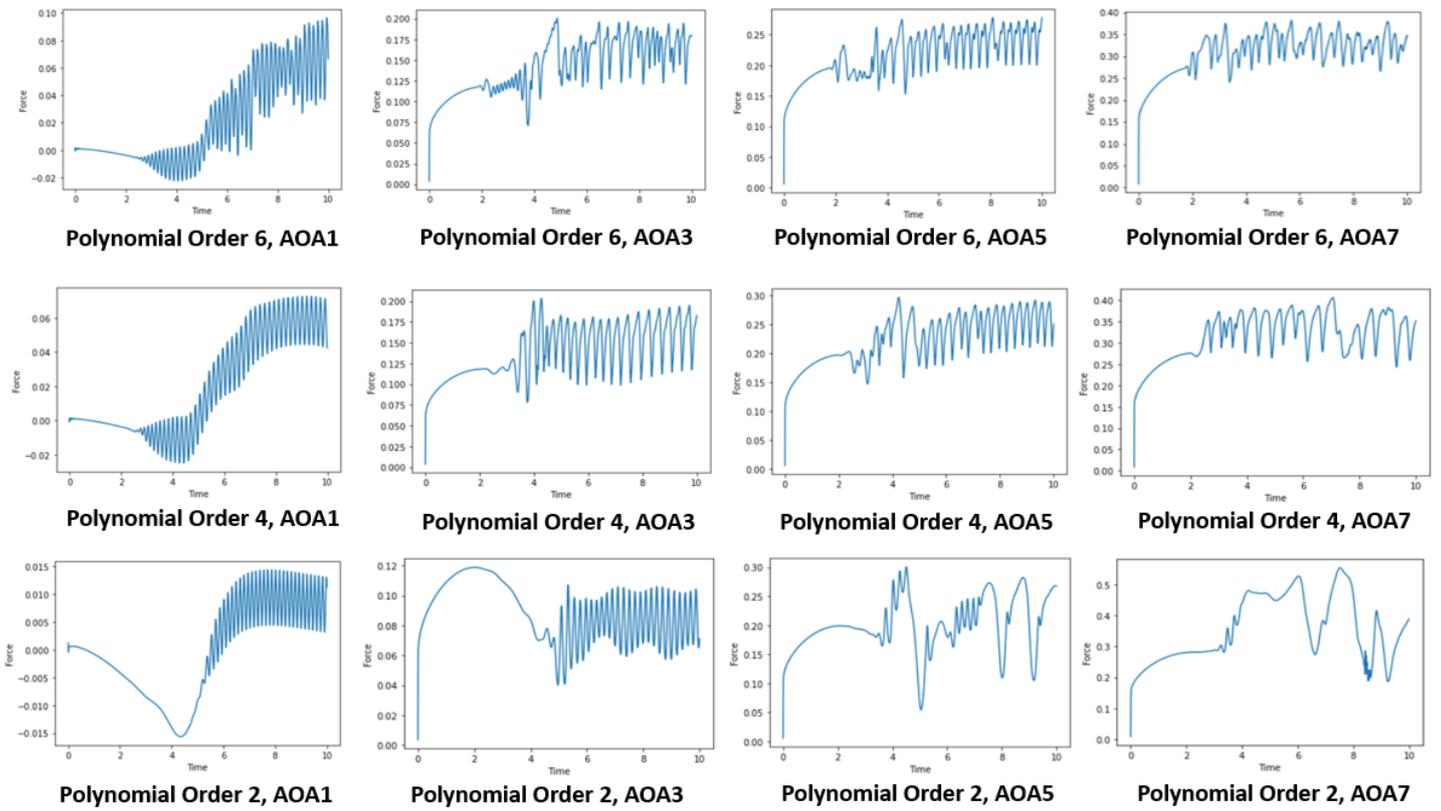

Figure B.3: Force results for Polynomial Order 6, 4 and 2 for different angle of attack = 1°, 3°, 5° and 7° respectively for 1 million iterations

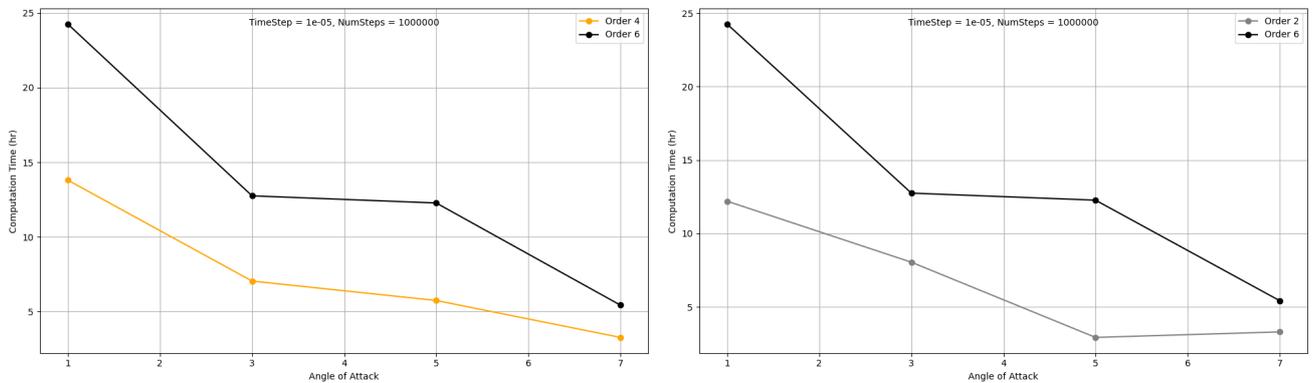

Figure B.4: Computation time comparison plot for different polynomial order with respect to time

# Appendix C

# Machine Learning

Machine Learning (ML) is a branch of artificial intelligence (AI) that concentrates on creating algorithms and models capable of enabling computers to acquire knowledge and formulate predictions or decisions grounded in data, all without the need for explicit programming. It can be classified into several major types:

***Supervised Learning:*** Supervised learning is a machine learning task in which the algorithm is trained on labeled data. This means that each data point is associated with a target or outcome. The goal of supervised learning is to learn a mapping from input to output and is suitable for classification and regression tasks.

***Unsupervised Learning:*** Unsupervised learning is a machine learning task in which the algorithm is trained on unlabeled data. The goal of unsupervised learning is to discover inherent patterns in the data, group similar data points, or reduce the dimensionality of the data without any prior knowledge about the labels.

***Reinforcement Learning:*** Reinforcement learning is a category of machine learning wherein an agent is trained to make decisions in an environment with the aim of maximizing a reward signal. This approach is frequently applied in situations where there is a need to learn a sequence of actions to accomplish a specific objective.

***Deep Learning:*** Deep Learning, which falls within the domain of Machine Learning (ML), is characterized by its emphasis on neural networks comprising numerous layers, commonly known as deep neural networks.

### C.0.1 Hyperparameter Optimization

There are many hyperparameters that can be optimized for DNNs. Here is a detailed description for hyperparameter:

**Batch size:** The batch size is the number of samples used in each forward and backward pass. It affects the model's convergence speed and memory usage. A larger batch size leads to faster convergence, but it also requires more memory.

**Number of layers:** The number of layers in a neural network is a hyperparameter that can affect the model's performance. A deeper network learn more complex relationships, but it is also more difficult to train.

**Number of neurons in each layer:** The number of neurons in each layer is a hyperparameter that can affect the model's accuracy and computational complexity. A larger number of neurons improves the model's accuracy, but it also makes the model more computationally expensive.

*Activation functions:* Activation functions are mathematical operations that are used to transform the output of each layer in a neural network.

**Weight initialization:** Weight initialization is the process of setting the initial values of the weights in a neural network. Different initialization methods can have different effects on the model's performance.

There are several hyperparameter optimization techniques for finding the best hyperparameters for the Deep Neural Network:

**Bayesian Optimization:** Bayesian optimization is a hyperparameter optimization technique that uses probabilistic models to guide the search for optimal hyperparameters efficiently.

**Genetic Algorithms:** Genetic algorithms are a hyperparameter optimization technique that employs the principles of natural selection to evolve a population of hyperparameter sets over multiple iterations. The best-performing individuals are selected and used to create new generations of hyperparameter sets.

**Gradient-Based Optimization:** Gradient-based optimization is a hyperparameter optimization technique that treats hyperparameters as continuous variables and optimizes them using gradient-based techniques. This approach is suitable for differentiable hyperparameters, such as learning rates.

### C.0.2 Activation functions

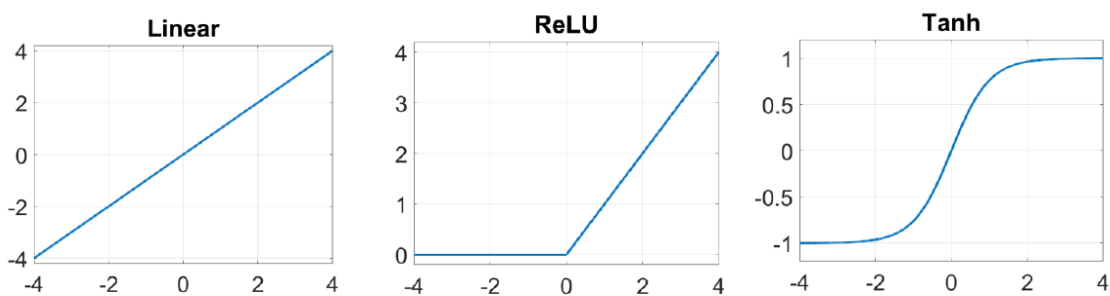

Figure C.1: Overview of activation functions

### C.0.3 Overfitting

Overfitting occurs when the model not only captures the overall patterns within the training data but also begins to tailor itself to the particular idiosyncrasies of individual training instances, including noise and other irregularities.

# Appendix D

# Adaptive Sampling

## IV. Adaptive Sampling

Kriging with adaptive sampling is commonly used to update the surrogate [26] [17] [8]. Infill criteria can take on many forms, such as expected improvement, goal seeking via the ln-likelihood, and infill via the model uncertainty [9]. Criteria that involve expected improvement or goal seeking are best used in cases of optimisation where there is a search for a global minimum. Updating the surrogate via the model uncertainty is better suited to the problem at hand of non-smooth stochastic outputs as it can be modified to include gradient information from the surrogate as will be shown below.

- Criterion 1, [8]

$$\text{Crit}(\boldsymbol{\xi}) = \text{PDF}(\boldsymbol{\xi})\hat{s}(\boldsymbol{\xi}) \qquad (1)$$

Figure D.1: Mathematical formulation for Adaptive Sampling [15]

The first criterion comes from [8]. It is based on the model uncertainty, $\hat{s}(\boldsymbol{\xi})$, augmented by the PDF($\boldsymbol{\xi}$), which is used to achieve faster stochastic convergence [26]. As it is based on the uncertainty and the uncertainty is based on the distance between adjacent sample points, this criterion is expected to evenly sample the stochastic space, particularly under uniform input uncertainty. Additionally, when using the uncertainty, no new points will be added at existing points, as at existing points $\hat{s}(\boldsymbol{\xi}) = 0$.

- Criterion 2, [26],[17]

$$\text{Crit}(\boldsymbol{\xi}) = \text{PDF}(\boldsymbol{\xi})\hat{s}(\boldsymbol{\xi})\left|\frac{\partial \hat{f}(\boldsymbol{\xi})}{\partial \boldsymbol{\xi}}\right|\Delta\boldsymbol{\xi} \qquad (2)$$

Criterion 2 is a modification of criteria from [26] and [17]. It uses the first derivative of the predictor, $\hat{f}(\boldsymbol{\xi})$, in addition to the uncertainty. The derivative is meant to find new sample points in the area of sharp gradients. The derivative of $\hat{f}(\boldsymbol{\xi})$ for kriging is given by

$$\frac{\partial \hat{f}(\boldsymbol{\xi})}{\partial \boldsymbol{\xi}} = \frac{\partial \boldsymbol{r}(\boldsymbol{\xi})}{\partial \boldsymbol{\xi}}^T \boldsymbol{R}^{-1}(\boldsymbol{f} - \boldsymbol{1}\mu) \qquad (3)$$

and the derivative of $\hat{f}_e(\boldsymbol{\xi})$ for co-kriging is given by

$$\frac{\partial \hat{f}_e(\boldsymbol{\xi})}{\partial \boldsymbol{\xi}} = \frac{\partial \boldsymbol{c}(\boldsymbol{\xi})}{\partial \boldsymbol{\xi}}^T \boldsymbol{C}^{-1}(\boldsymbol{f} - \boldsymbol{1}\mu). \qquad (4)$$

The derivative of $\boldsymbol{r}$ can be calculated analytically from the correlation function

$$\left[\frac{\partial \boldsymbol{r}(\boldsymbol{\xi})}{\partial \boldsymbol{\xi}}\right]_{i,j} = k'(\xi_j, \xi_j^{(i)}) \times \prod_{\substack{k=1 \\ k \neq j}}^{M} k(\xi_k, \xi_k^{(i)}) \qquad (5)$$

and the derivative of $\boldsymbol{c}$ can be calculated using the definition of the derivative of $\boldsymbol{r}$ above

$$\begin{aligned}
\left[\frac{\partial \boldsymbol{c}(\boldsymbol{\xi})}{\partial \boldsymbol{\xi}}\right]_{m,j} &= \hat{\rho}\hat{\sigma}_c^2 \left[\frac{\partial \boldsymbol{r}(\boldsymbol{\xi})}{\partial \boldsymbol{\xi}}\right]_{m,j} & m = 1, 2, \ldots N_c \\
\left[\frac{\partial \boldsymbol{c}(\boldsymbol{\xi})}{\partial \boldsymbol{\xi}}\right]_{n,j} &= \hat{\rho}^2\hat{\sigma}_c^2 \left[\frac{\partial \boldsymbol{r}(\boldsymbol{\xi})}{\partial \boldsymbol{\xi}}\right]_{n,j} + \hat{\sigma}_d^2 \left[\frac{\partial \boldsymbol{r}(\boldsymbol{\xi})}{\partial \boldsymbol{\xi}}\right]_{n,j} & n = 1, 2, \ldots N_e,
\end{aligned} \qquad (6)$$

where $m$ is the index of the cheap data and $n$ is the index of the expensive data, since $\boldsymbol{c}$ is comprised of correlation of both data sets.

For the universal cubic correlation function, the term $k'(\xi_j, \xi_j^{(i)})$ in equation 5 is defined as

$$k'(\xi_j, \xi_j^{(i)}) = -\frac{6(1-\rho_j)}{2+\gamma_j}h_j + \frac{3(1-\rho_j)(1-\gamma_j)}{2+\gamma_j}h_j|h_j|. \qquad (7)$$

Figure D.2: Mathematical formulation for Adaptive Sampling [15]

In Criterion 2, the term $\Delta\xi$ is defined as [26]

$$\Delta\xi = \min_{i=1,2,\ldots,N} \left|\xi - \xi^{(i)}\right|. \qquad (8)$$

This term is intended to help the predictor find samples in areas in the stochastic space where samples are sparse. It acts similarly to the uncertainty and helps to provide a balance between the uncertainty and gradient terms. One of the problems with this criterion is that it will not find samples where $\frac{\partial \hat{f}(\xi)}{\partial \xi} = 0$. Additionally, use of the first derivative can result in overshoots of the non-smooth behaviour as will be shown in §V A.

- Criterion 3

$$\text{Crit}(\xi) = \text{PDF}(\xi)\hat{s}(\xi) \left|\frac{\partial \hat{f}(\xi)}{\partial \xi} \frac{\partial^2 \hat{f}(\xi)}{\partial \xi^2}\right| \Delta\xi \qquad (9)$$

The adaptive sampling criteria proposed in this work are presented in Criterion 3, 4, 5(a), and 5(b). These new criteria use the second derivative of the kriging predictor in addition to the first derivative. The advantage of using the second derivative coupled with the first derivative is that the second derivative helps to capture the inflection features of the output. This is especially beneficial in the case of a discontinuity when the sharp gradient is centred on the peak of a normal PDF. In this case, Criterion 2 will have difficulty finding features of the curve outside of the region of the sharp gradient.

The second derivative of $\hat{f}(\xi)$ is given by

$$\frac{\partial^2 \hat{f}(\xi)}{\partial \xi^2} = \frac{\partial^2 r(\xi)}{\partial \xi^2}^T R^{-1}(f - 1\mu) \qquad (10)$$

and the second derivative of the co-kriging predictor, $\hat{f}_c(\xi)$, is given by

$$\frac{\partial^2 \hat{f}_c(\xi)}{\partial \xi^2} = \frac{\partial^2 c(\xi)}{\partial \xi^2}^T C^{-1}(f - 1\mu). \qquad (11)$$

The second derivative of $r$ is defined as

$$\left[\frac{\partial^2 r(\xi)}{\partial \xi}\right]_{i,j} = k''(\xi_j, \xi_j^{(i)}) \times \prod_{\substack{k=1 \\ k \neq j}}^{M} k(\xi_k, \xi_k^{(i)})$$

$$+ \sum_{\substack{l=1 \\ l \neq j}}^{M} k'(\xi_j, \xi_j^{(i)}) \times k'(\xi_l, \xi_l^{(i)}) \times \prod_{\substack{k=1 \\ k \neq j,l}}^{M} k(\xi_k, \xi_k^{(i)}) \qquad (12)$$

Figure D.3: Mathematical formulation for Adaptive Sampling [15]

and the second derivative of $c$ is defined similarly to equation 6

$$\left[\frac{\partial^2 c(\xi)}{\partial \xi^2}\right]_{m,j} = \hat{\rho}\hat{\sigma}_c^2 \left[\frac{\partial^2 r(\xi)}{\partial \xi}\right]_{m,j} \qquad m = 1, 2, \ldots N_c$$

$$\left[\frac{\partial^2 c(\xi)}{\partial \xi^2}\right]_{n,j} = \hat{\rho}^2\hat{\sigma}_c^2 \left[\frac{\partial^2 r(\xi)}{\partial \xi}\right]_{n,j} + \hat{\sigma}_d^2 \left[\frac{\partial^2 r(\xi)}{\partial \xi^2}\right]_{n,j} \qquad n = 1, 2, \ldots N_e, \tag{13}$$

while Criterion 3 is updated with the second derivative of the predictor, it will still be limited in the same way as Criterion 2 in that no new samples will be found where the first and second derivatives of the predictor are equal to zero.

- Criterion 4

Criterion 4 also uses the second derivative of the predictor, but with the second and first derivatives being combined via addition as opposed to multiplication. This criterion is expected to add points similarly to Criterion 3, however, without the restriction of no new points being added when $\frac{\partial \hat{f}(\xi)}{\partial \xi} = 0$.

$$\text{Crit}(\xi) = \text{PDF}(\xi)\hat{s}(\xi)\left|\frac{\partial \hat{f}(\xi)}{\partial \xi} + \frac{\partial^2 \hat{f}(\xi)}{\partial \xi^2}\right|\Delta\xi \tag{14}$$

- Criterion 5(a)

$$\text{Crit}(\xi) = \text{PDF}(\xi)\hat{s}(\xi)\left(\left|\frac{\partial \hat{f}(\xi)}{\partial \xi} + \frac{\partial^2 \hat{f}(\xi)}{\partial \xi^2}\right|\Delta\xi + D_{f_G}^{UC}\right) \tag{15}$$

- Criterion 5(b)

$$\text{Crit}(\xi) = \text{PDF}(\xi)\hat{s}(\xi)\left(\left|\frac{\partial \hat{f}(\xi)}{\partial \xi} + \frac{\partial^2 \hat{f}(\xi)}{\partial \xi^2}\right|\Delta\xi + D_{f_C}^{UC}\right) \tag{16}$$

where [26]

$$D_{f_G}^{UC} = \left|\hat{f}_{UC} - \hat{f}_G\right| \tag{17}$$

$$D_{f_G}^{UC} = \left|\hat{f}_{UC} - \hat{f}_C\right|. \tag{18}$$

Equations 17 and 18 are defined as the difference between the kriging predictor constructed with the universal cubic correlation ($\hat{f}_{UC}$) function and the Gaussian correlation ($\hat{f}_G$) and cubic correlation ($\hat{f}_C$) functions. These terms provide an additional error estimate as the cubic and Gaussian correlation function behave differently from the universal cubic in regions

Figure D.4: Mathematical formulation for Adaptive Sampling [15]